\documentclass{article}

\usepackage[preprint]{neurips_2026}

\usepackage[utf8]{inputenc} 
\usepackage[T1]{fontenc}    
\usepackage{hyperref}       
\usepackage{url}            
\usepackage{booktabs}       
\usepackage{amsfonts}       
\usepackage{nicefrac}       
\usepackage{microtype}      
\usepackage{xcolor}         
\usepackage{xspace}
\usepackage{inconsolata}
\usepackage{listings}
\usepackage{multirow}
\usepackage{fontawesome5}
\usepackage[most]{tcolorbox}

\tcbset{%
    enhanced,
    coltitle=black, 
    detach title,
    left=10mm,
    overlay={
        \node[minimum width=1cm,xshift=5mm] at (frame.west) {\tcbtitle};
    },
    boxrule=1pt
}

\usepackage{graphicx}
\usepackage{caption}
\usepackage{tcolorbox}
\usepackage{amsmath}
\usepackage{subcaption}
\usepackage{float}
\floatstyle{plain}
\newfloat{listing}{tbp}{lol}
\floatname{listing}{Listing}
\usepackage{multicol}

\newcommand{\fxb}[0]{\emph{fx}Bench\xspace}
\newcommand{\accuracy}[0]{\textsc{acc}\xspace}
\newcommand{\accuracyUpper}[0]{\textsc{acc}$^\dagger$\xspace}
\newcommand{\rangeOverlap}[0]{\textsc{ovl}\xspace}

\definecolor{Set1Red}{RGB}{228, 26, 28}
\definecolor{Set1Blue}{RGB}{55, 126, 184}
\definecolor{Set1Green}{RGB}{77, 175, 74}
\definecolor{codebg}{rgb}{0.95,0.95,0.95}
\lstdefinelanguage{excel}{
    showstringspaces=false,
    breaklines=true,
}
\lstdefinestyle{excel}
{
    language=excel,
    basicstyle=\small\ttfamily,
    string=[d]{"},
    stringstyle=\color{Set1Blue},
    keywordstyle=\color{blue},
    morekeywords={select,index},
    showspaces=false,
    showstringspaces=false,
    backgroundcolor = \color{codebg},
    frame=tb,
    rulecolor=\color{codebg},
    xleftmargin=1em,
    framexleftmargin=1em,
    columns=flexible
}

 \definecolor{ghBg}{HTML}{F6F8FA}
 \definecolor{ghBorder}{HTML}{D0D7DE}
 \definecolor{ghText}{HTML}{1F2328}
 
 \newcommand{\gcode}[1]{%
   \tcbox[on line, colback=ghBg, colframe=ghBorder,
          boxrule=0.4pt, arc=2pt, boxsep=1pt,
          left=2pt, right=2pt, top=0pt, bottom=0pt]%
         {\ttfamily\small\color{ghText}#1}%
 }
 
 \newtcblisting{gblock}{
   listing only,
   listing options={
     basicstyle=\ttfamily\small\color{ghText},
     breaklines=true,
     breakatwhitespace=true,
     columns=fullflexible,
     keepspaces=true,
     showstringspaces=false,
     upquote=true,
     aboveskip=0pt, belowskip=0pt,
   },
   colback=ghBg, colframe=ghBorder,
   boxrule=0.5pt, arc=3pt,
   left=6pt, right=6pt, top=4pt, bottom=4pt,
   enhanced, breakable,
 }

 \newtcblisting{gblocksmall}{
   listing only,
   listing options={
     basicstyle=\ttfamily\tiny\color{ghText},
     breaklines=true,
     breakatwhitespace=true,
     breakindent=0pt,            
     breakautoindent=false,      
     columns=fullflexible, keepspaces=true,
     showstringspaces=false, upquote=true,
     escapeinside={(*@}{@*)},
     aboveskip=0pt, belowskip=0pt,
   },
   colback=ghBg, colframe=ghBorder,
   boxrule=0.5pt, arc=3pt,
   left=6pt, right=6pt, top=4pt, bottom=4pt,
   enhanced,                      
 }

\title{\emph{fx}Bench: Evaluating and Understanding Formula Suggestions in Spreadsheets}

\author{
    Sanket Mhatre\thanks{Work completed while at Microsoft.} \\
    Proximal\\
    \texttt{sanket@proximal.so} \\
    \And 
    Sumit Gulwani\\
    Microsoft, USA\\
    \texttt{sumitg@microsoft.com} \\
    \And 
    Vu Le\\
    Microsoft, USA\\
    \texttt{levu@microsoft.com} \\
    \And 
    Yasharth Bajpai\\
    Microsoft, India\\
    \texttt{ybajpai@microsoft.com} \\
    \And 
    Gust Verbruggen\\
    Microsoft, Belgium\\
    \texttt{gverbruggen@microsoft.com} \\
}

\begin{document}

\maketitle

\begin{abstract}
Spreadsheets are among the most widely used tools for computations, and formulas are their core building blocks.
Besides agentic spreadsheet tools, proactive formula suggestions enable significant productivity gains and are now available in popular spreadsheet software (Microsoft Excel and Google Sheets).
Such suggestions pose interesting challenges: the right context must be collected and represented for a machine learning model, which must then be able to write a correct formula, in a way that is fast and cheap enough to trigger on every \texttt{=} (or subsequent keystrokes for auto-completion) without taking users out of their editing flow.
Surprisingly, there is no public benchmark for formula suggestions on real spreadsheets.
In this paper, we therefore (1) introduce \fxb as a manually curated benchmark of 503 formula suggestion tasks based on Sheetpedia, (2) break down the formula suggestion problem into four capabilities---gather context, understand context, understand intent, write formula---that a formula suggestion system requires, and (3) evaluate existing and new approaches to understand bottlenecks in these capabilities that pose interesting research directions.
Specifically, we annotate each benchmark with the cells that describe intent to determine the theoretical performance limit of a given context selection strategy, we evaluate performance of small and large models on different representations of context---including a dense format with high theoretical coverage at low token counts---and we fine-tune smaller language models to show that we can learn to understand such dense formats.
Additionally, we perform quantitative and qualitative analysis on failure modes to understand bottlenecks in the formula writing capability---the best configuration only achieved 58.8\% correct suggestions.

\end{abstract}
\begin{figure}[htb]
    \centering
    \caption{The formula suggestion workflow. Given a spreadsheet and a target cell (left), a pre-processing step extracts and serializes the relevant context (middle), which is passed to an LLM that predicts the formula (right).}
    \includegraphics[width=\linewidth]{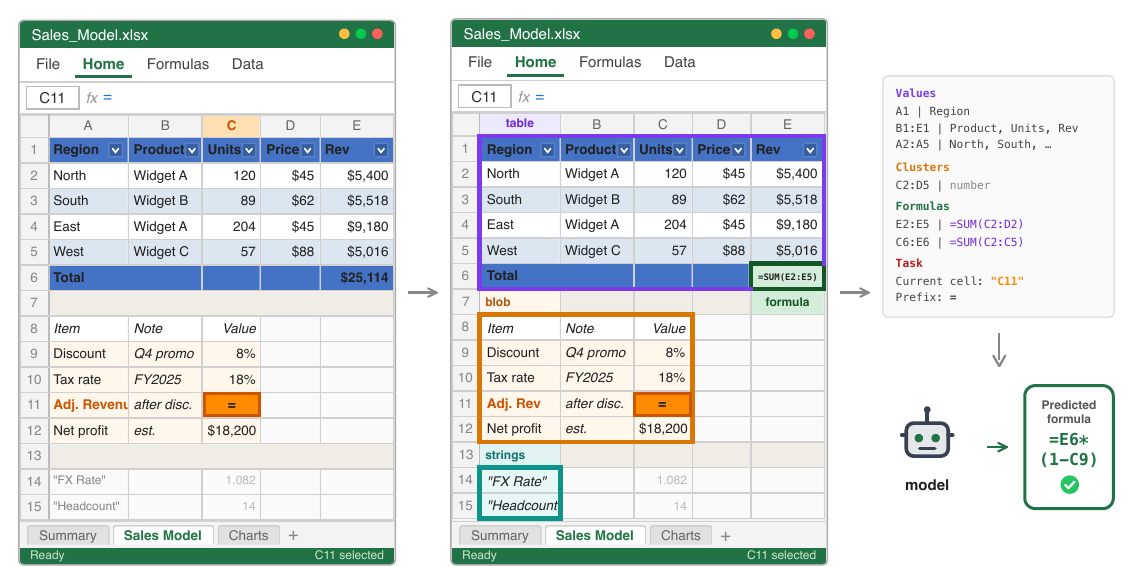}
    \label{fig:fc}
\end{figure}

\section{Introduction}


Spreadsheets are among the most widely used computational tools in the world, and formulas are the computational backbone of every spreadsheet.
Yet, writing such formulas is still tedious and prone to errors for end-users \citep{automatingstringprocessing,Hermans2012Understandability}.
Modern spreadsheet assistants\footnote{Copilot in Microsoft Excel, Gemini in Google Sheets, or external tools like Shortcut, Claude for Excel, ChatGPT for Excel, and more.} aim to resolve this by allowing users to describe specifications in natural language.
Often, however, the required context is already there, and describing it requires more keystrokes or precise language than simply writing the formula.
Excel and Google Sheets therefore both expose \emph{formula suggestions} that trigger when users type the \texttt{=} sign, indicating that they are about to write a formula.

SpreadsheetCoder \citep{chen2021spreadsheetcoder} introduced this task---suggesting a formula from the surrounding context---as a research problem, but (1) did not introduce a public benchmark, (2) relied on a fixed and local context that limits applicability, (3) used embeddings (BERT, $\approx 40$M parameters, 512 tokens context) to predict formulas using LSTMs, which limited capabilities compared to modern decoder models.
In this paper, we tackle each of these limitations with the following main contributions.

\textbf{We present \fxb, the first benchmark for formula suggestions}, which consists of of 503 manually verified cases spanning 147 Excel formula types across 441 real workbooks from the Sheetpedia corpus \citep{tian2025sheetpedia}.
Each case consists of a workbook, a target cell, and either the expected formula (454 cases) or an abstention indicator where we expect no formula because the intent is not clearly specified (49 cases).
An example of the formula suggestion problem and language model solution is shown in Figure~\ref{fig:fc}.

\textbf{We break down formula suggestions in four capabilities} by annotating each problem with the spreadsheet cells that describe the intent for that formula (and release those annotations), as well as a natural language description.
This allows us to perform isolated experiments on context gathering, representation understanding, intent understanding, and formula writing.


\textbf{We evaluate pre-trained and fine-tuned models on \fxb} to understand key properties of the formula suggestion problem, including \textbf{(1)} the effect of context gathering (higher theoretical performance, lower actual performance) and representation (9.1\% difference between best and worst) using inference-only models, and \textbf{(2)} the effect of supervised fine-tuning on smaller open models (Qwen3, \cite{qwen3}), showing that the dense spreadsheet representation can be taught (pass@1 $+$11.1\%, pass@3 $+$18.2\%, best@8 $+$23.5\% on the 8B model), that gains begin to saturate within Sheetpedia, and that fine-tuned models often produce a correct formula in their top-$k$ candidates but struggle to rank it first (best@8 exceed pass@1 by $+$16.8\% for 8B and $+$20.7\% pp on 14B).

\section{Related Work}

Spreadsheets have long been a setting for end-user programming and intelligent assistance.
Earlier programming-by-example for spreadsheets synthesized transformations from user-provided examples \cite{automatingstringprocessing}.
Subsequent work has studied spreadsheet formula prediction more directly: SpreadsheetCoder \citep{chen2021spreadsheetcoder} predicts formulas from semi-structured spreadsheet context; Fortap \citep{cheng2022fortap} uses formulas for numerical reasoning aware table pretraining and evaluates formula prediction; HermEs \citep{he-etal-2023-hermes} improves formula prediction through hierarchical formula expansion and an interactive completion interface; and Auto-Formula \citep{chen2024auto} recommends formulas by retrieving and adapting formulas from similar spreadsheets.
None of these release a curated benchmark or support generic and novel formulas based on surrounding intent.
Other formula oriented tasks change the available specification: \textsc{flame} \citep{joshi2024flame} models formula syntax and prefix completion, NL2Formula \citep{zhao2024nl2formula} uses explicit natural-language intent, and FoRepBench \citep{singha2025benchmark} repairs faulty formulas. 

Recent work also studies broader spreadsheet agents and table representations for LLMs.
SpreadsheetBench \citep{ma2024spreadsheetbench}, SheetCopilot \citep{li2023sheetcopilot}, Finch \citep{dong2025finch}, and InstructExcel \citep{payan2023instructexcel} evaluate spreadsheet manipulation from explicit natural language tasks, with tools and multi-turn interaction.
They measure broad workflow completion rather the inline authoring moment.
Separately, table and spreadsheet representation work shows that serialization, sampling, packing, and structure-aware encoding strongly affect LLM performance \citep{sui2024table, sui2024tap4llm, dong2024spreadsheetllm}.
Table-specialized models have also improved reasoning through pretraining, instruction tuning, learned table encoders, and parameter-efficient adaptation \citep{wang2021tuta, yin2020tabert, herzig2020tapas, he2025tablelora, zhang2024tablellama, xing2025table, zhang2025tablellm, liu2021tapex}.
Sheetpedia \citep{tian2025sheetpedia} provides a corpus of 300K real-world spreadsheets, and serves as the source corpus for \fxb.

\section{\fxb}

\fxb is a benchmark of 503 manually verified formula completion records spanning 147 unique Excel formula types across 441 distinct spreadsheets.
We use Sheetpedia---a collection of 300K spreadsheets compiled from enterprise email archives, public web crawls, and Excel forums, covering domains like financial models, inventory tracking, and academic data---as the source of spreadsheets \citep{tian2025sheetpedia}.
Each record consists of (1) a version of the workbook with potential leaks removed, (2) the ground-truth formula authored by the original user, and (3) the target cell.


\begin{figure}
    \centering
    \caption{The \fxb curation pipeline.}
    \includegraphics[width=0.92\linewidth]{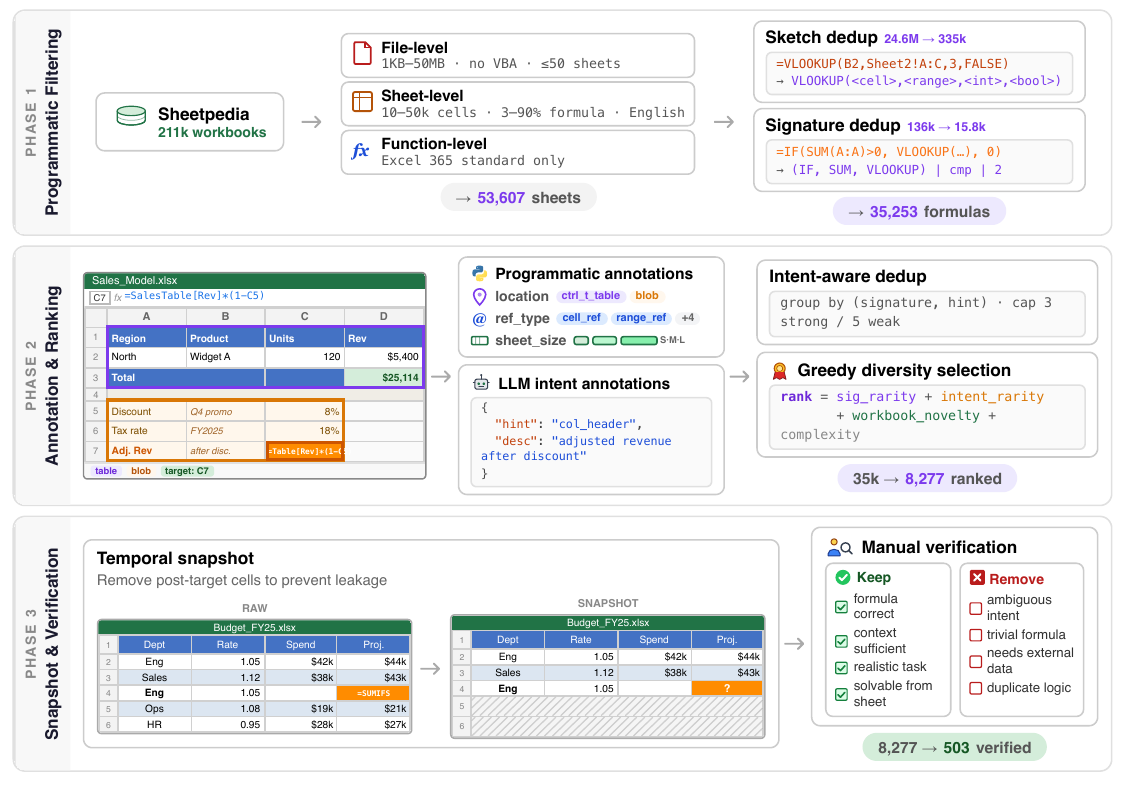}
    \label{fig:curation}
\end{figure}

\subsection{Construction}
\label{main:fxbench-construction}

An overview of the curation process is shown in Figure~\ref{fig:curation}.
We briefly highlight the three main steps in this section.
Full details, examples, and LLM prompts are provided in Appendix~\ref{app:fxbench}.

\paragraph{Filtering}

We filter \textbf{files} that are encrypted, corrupted, or $\notin [1 \text{KB}, 50\text{MB}]$, (2) files with \textbf{sheets} with cells $\notin [10, 50\text{K}]$ cells, formula ratio $\notin [3\%, 90\%]$, not English, has any external link, and (3) files with \textbf{functions} that are not in the Excel 365 standard library.
This reduces 211K workbooks down to 53.6K, yielding $\approx 24.6$M formulas.

To control redundancy, we apply two levels of de-duplication. \textbf{Sketch de-duplication} replaces literal arguments with typed placeholders (\lstinline[style=excel]|SUM(<range>)|) and keeps one record per (workbook, sketch) pair, retaining 335K formulas across 136K unique sketches. \textbf{Signature de-duplication} further abstracts each sketch into a tuple of (ordered function list, operator group set, nesting depth), where operators are binned into \textit{arithmetic}, \textit{comparison}, and \textit{concatenation}. For example, the sketches \lstinline[style=excel]|SUM(<range>)|, \lstinline[style=excel]|SUM(<int>)|, and \lstinline[style=excel]|SUM(<cell>, <cell>, <cell>)| all share the signature \lstinline[style=excel]{(SUM) | | 1} (one function, no operators, depth one) but differ in argument structure.
This collapses 136K sketches into 15,845 signatures.
23,963 representative sketches are selected per signature---maximizing frequency, complexity, and argument-style diversity---which are then sampled at up to 5 formulas per sketch, yielding 35,253 candidates.

\paragraph{Annotation and Ranking}
We annotate each candidate and select a ranked subset for manual review.
Symbolic annotations are target cell location relative to tables and blobs, reference types, and sheet size.
An LLM annotates each candidate with the dominant contextual intent available in the worksheet: column header, row label, adjacent formula, adjacent value, combination, or no hint (based on local context + local context of references).
These categories describe what cues are present, not what models use, and are only used to curate a balanced distribution across hint types.
De-duplication based on intent and greedy diversity selection (Appendix \ref{app:benchmark-phase-2}) reduce 35,253 candidates to 8,277 spanning 296 functions.

\paragraph{Manual Curation and Verification}
Two challenges remain.
To address \textbf{leakage}---cells below or to the right of the target often follow obvious patterns a model can copy rather than infer---we build the state of the workbook as it plausibly existed at the moment the user was about to write the formula, which is also the state an inline formula completion system observes at inference time.
A two-pass LLM pipeline segments the sheet into logical regions and their ordering, and strips regions added after the region with the target cell (Appendix \ref{app:temporal}).
Within the target region, we remove formulas with the same template below and to the right of the target formula.
To address \textbf{solvability}, we inspect both the original workbook and the snapshot, and retain a record only if the snapshot preserves enough context to infer the target formula while removing post-target leakage.
Each snapshot is reviewed for formula correctness, context sufficiency, task realism, and solvability without external domain knowledge.
A subset of 49 records are flagged as \emph{abstain}: target cells where no formula should reasonably be written, testing a model's ability to withhold prediction when context is insufficient. 
The 8,277 candidates reduce to the final 503 verified records spanning 147 formula types: 145 Excel functions, expression-only formulas and abstention cases.

\begin{figure}[htb]
\centering
\begin{minipage}[t]{0.42\textwidth}
\vspace{0pt}
\captionof{table}{Properties of \fxb.}
\label{tab:distributions}
\scriptsize
\begin{tabular}{lr}
\toprule
\textbf{Feature} & \textbf{\% of Records} \\
\midrule
\multicolumn{2}{l}{\textit{Target cell location}} \\
\quad Inside structured region (table cell)              & 62.0\% \\
\quad Adjacent to structured region                      & 33.4\% \\
\quad Unstructured                                       & 4.6\% \\
\midrule
\multicolumn{2}{l}{\textit{Formula reference types\textsuperscript{$\dagger$}}} \\
\quad Range reference (\texttt{A1:A10})                  & 59.2\% \\
\quad Cell reference (\texttt{A1})                       & 58.4\% \\
\quad Absolute reference (\texttt{\$A\$1})               & 47.5\% \\
\quad Cross-sheet reference                              & 8.5\% \\
\quad Whole row/column (\texttt{A:A})                    & 8.3\% \\
\quad Named reference                                    & 4.8\% \\
\midrule
\multicolumn{2}{l}{\textit{Intent hint source (strongest contextual clue)}} \\
\quad Column header                               & 44.3\% \\
\quad Row label                                   & 31.6\% \\
\quad Adjacent formula                            & 10.5\% \\
\quad Combination                                 & 9.9\% \\
\quad No hint                                     & 2.0\% \\
\quad Adjacent values                             & 1.6\% \\
\bottomrule
\end{tabular}\\
{\scriptsize \textsuperscript{$\dagger$} Not mutually exclusive}
\end{minipage}
\hfill
 \begin{minipage}[t]{0.56\textwidth}
 \vspace{0pt}
 \centering
 
 \captionof{figure}{\fxb distribution by function category, nesting depth, and formula length}
 \includegraphics[width=0.95\linewidth]{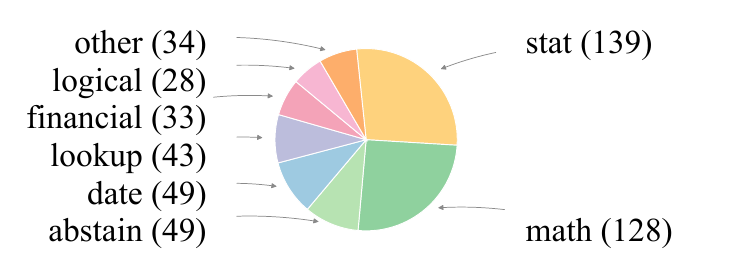}
 \label{fig:function_category_share}
 
 \vspace{0pt}
 

 \centering
 
 \scriptsize
 \begin{tabular}{lr@{\hspace{12pt}}lr}
 \toprule
 \textbf{Depth} & \textbf{Count} & \textbf{Length} & \textbf{Count} \\
 \midrule
 0  & 46  & 0--25    & 105 \\
 1  & 144 & 25--50   & 161 \\
 2  & 113 & 50--75   & 63  \\
 3  & 94  & 75--100  & 36  \\
 4  & 49  & 100--200 & 63  \\
 5  & 26  & 200--300 & 15  \\
 6+ & 31  & 300+     & 11  \\
 \bottomrule
 \end{tabular}

 \end{minipage}
\end{figure}

\subsection{Evaluation}

\paragraph{Execution-based Evaluation}

We evaluate by execution, not string match, to avoid false negatives for equivalent formulas such \lstinline[style=excel]|SUM(A1:A10)| and \lstinline[style=excel]|A1+A2+...+A10|. 
We write the predicted and ground-truth formulas into the target cell and compare their formatted displayed values.\footnote{Through \texttt{xlwings}: \url{www.xlwings.org}.}

\paragraph{Metrics}

\textbf{TP} (predicted and expected display values match), \textbf{FP} (prediction parsed but produced wrong value or error value), \textbf{TN} (abstention expected and model abstained), \textbf{FN} (formula expected but model abstained).
We report \textbf{pass@k accuracy} as the primary metric \citep{chen2021evaluating}.
\section{Decomposing Formula Suggestions}

We break down the formula suggestion problem into four main capabilities that are needed: \emph{context gathering}, \emph{context understanding}, \emph{intent understanding} and \emph{formula writing}.
In the following three subsections---intent and formula are combined---we perform experiments that evaluate properties of these capabilities to determine core challenges for future work on this problem.

\begin{figure}
    \centering
    \caption{Minisheet representation. Blob detection finds contiguous non-empty regions; edge extraction keeps boundary cells; semantic grouping emits a compact view of values, numeric clusters, formulas and tables sections.}
    \includegraphics[width=\linewidth]{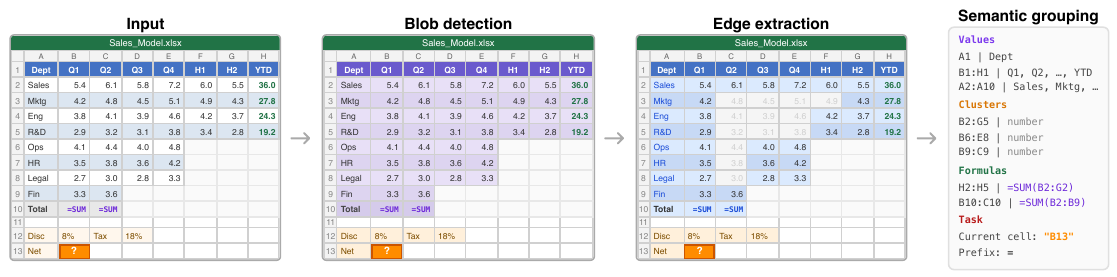}
    \label{fig:minisheet-pipeline}
\end{figure}

\subsection{Context Gathering}
\label{ssec:context}

Given a spreadsheet and a target cell, the first challenge is determining the right context: the cells that provide the relevant intent.
There is a trade-off between the size of the context and the theoretical performance of formula suggestions under that context.
Motivations for keeping the context small are speed, cost, and the needle-in-a-haystack effect \citep{gao2026u,li2025longcontext}.

To analyze context gathering without the effects of context understanding and formula writing, we annotate each problem with the cells that describe its intent.
We build a small, local context of all cells within 4 rows and columns of the target cell and each corner of a range (\lstinline[style=excel]|B6:F23| $\rightarrow$ (\texttt{\small B6}, \texttt{\small F6}, \texttt{\small B23}, \texttt{\small F23})) and use an LLM (GPT 5.2) to select the cells that describe the direct intent (like ``\emph{Total}'') and data description (like ``\emph{Cats}''). Full Details are provided in Appendix \ref{app:intent-annotation}.

Given any serialization (\emph{prompt}) of a context gatherer, we expect (1) all direct intent to be printed (address inferable, content visible) and (2) all data intent and data references to be mentioned---this is a heuristic \emph{upper limit} on the performance of an oracle agent (Refer Appendix \ref{app:theoretical-analysis}). 

We consider the following serializations:
\begin{itemize}
    \item \textbf{Full-sheet serialization.} Render each non-empty (address, value) pair in a sheet. Token cost grows linearly with sheet size, prohibitive for non-trivial spreadsheets.
    \item \textbf{Windowed Context.} Windowed approaches restrict serialization to a fixed region around the target cell. We evaluate two pipe-delimited variants: \textbf{Viewport} ($20$ rows, $10$ columns centered on the target cell) and \textbf{SpreadsheetCoder} \citep{chen2021spreadsheetcoder}, a tighter $10\times10$ window augmented with the detected table header row. Both are token-bounded but lose coverage when relevant cells fall outside the window.
    \item \textbf{SheetCompressor} \citep{dong2024spreadsheetllm} compresses spreadsheets through three sequential modules: structural-anchor extraction, inverted-index translation (merges cells with identical values into a value-keyed dictionary), and data-format aware aggregation into type-address tuples. String cells with arbitrary values preserve their exact text, and formulas are rendered individually with full text. The approach achieves full coverage with substantial token reduction on regular tabular layouts.\footnote{Neither SpreadsheetCoder nor SheetCompressor has an official public implementation; we re-implement both following the original papers. Reference code is released with \fxb.}
    \item We introduce \textbf{Minisheet}, a compact context representation for inline formula completion, based on simple ideas: (1) detect blobs on non-empty cells using flood fill, (2) print the $w$ boundary string cells of each blob, (3) collapse sibling formulas, (4) leverage the inverted-index translation from SheetCompressor to print every numeric range (clustered by format string), and (5) explicitly print information from structured tables. This is shown in Figure~\ref{fig:minisheet-pipeline}.
\end{itemize}

\begin{table}[htb]
\centering
\caption{Required-cell exposure coverage and token cost by context strategy on \fxb.}
\label{tab:context-evidence-metrics-short-v4}
\scriptsize
\setlength{\tabcolsep}{3pt}
\begin{tabular}{lrrrrrrr}
\toprule
\textbf{Context} & \textbf{Intent printed} & \textbf{Intent mentioned} & \textbf{Ref. printed} & \textbf{Ref. mentioned} & \textbf{Upper} & \textbf{Tok (Avg)} & \textbf{Tok (Median)} \\
\midrule
Pipe-delimited & 100.0\% & 100.0\% & 100.0\% & 100.0\% & 100.0\% & 19,984 & 3,128 \\
Viewport ($20 \times 10$) & 82.1\% & 82.1\% & 70.8\% & 70.8\% & 69.0\% & 2,603 & 1,913 \\
Viewport ($30 \times 20$) & 94.6\% & 94.6\% & 85.7\% & 85.7\% & 85.1\% & 3,806 & 2,574 \\
Viewport ($40 \times 30$) & 98.2\% & 98.2\% & 90.1\% & 90.1\% & 89.9\% & 4,533 & 2,881 \\
SpreadsheetCoder & 74.8\% & 74.8\% & 61.8\% & 61.8\% & 59.0\% & 1,806 & 1,364 \\
SheetCompressor & 90.5\% & 100.0\% & 43.5\% & 100.0\% & 94.6\% & 5,156 & 1,372 \\
\midrule
Minisheet ($w=1$) & 80.7\% & 91.8\% & 34.0\% & 90.1\% & 78.1\% & 1,736 & 673 \\
Minisheet ($w=2$) & 86.3\% & 98.6\% & 35.4\% & 96.6\% & 88.3\% & 1,798 & 725 \\
Minisheet ($w=3$) & 87.1\% & 99.4\% & 35.6\% & 98.4\% & 90.5\% & 1,817 & 732 \\
Minisheet ($w=4$) & 87.5\% & 99.8\% & 35.6\% & 99.0\% & 91.5\% & 1,842 & 738 \\
\bottomrule
\end{tabular}
\par\smallskip
\end{table}

Table~\ref{tab:context-evidence-metrics-short-v4} shows the how many of the required cells are mentioned and printed, as well as token statistics (\texttt{\small cl100k\_base}) for each method.
Figure~\ref{fig:token-comparison-full} shows how the median number of tokens grows in function of \# of non-empty cells for different approaches.

\paragraph{Local context is not sufficient}

Full-sheet serialization guarantees coverage but is prohibitively expensive---especially for large sheets (20K average versus 3.1K median).
Window-based methods are cheaper, but miss long-range dependencies: increasing the viewport from $20 \times 10$ to $40 \times 30$ leaves about $10\%$ of problems without all required evidence, while increasing median tokens by $1.5\times$.
The gap is larger for references than for intent, suggesting that local context often captures nearby labels but misses distant data ranges.
Minisheet and SpreadsheetCoder instead combine global summarization of potential data ranges with local structural anchors.

\paragraph{Mentioning unlocks dense representations}

By focusing on printing key string values and mentioning clustered numeric ranges, Minisheet is able to produce a very dense representation, losing at most $-$1.6\% theoretical coverage compared to the biggest viewport at 4$\times$ fewer tokens.
SheetCompressor also uses blob-based heuristics (based on tracking changes between subsequent rows and columns) and achieves high coverage, but also prints significantly more values ($+$8.1\% references) yielding a $\sim 1.9 \times$ increase in median \# of tokens.

\paragraph{Intent lives on the edges of blobs}
Increasing the border width $w$ in Minisheet yields diminishing returns for printing intent after $w=2$ ($-$1.2\% versus $w=4$ ) indicating that intent lives on these edges.
The 12.5\% gap to printing all intent happens due to semantic blobs being connected, which SheetCompressor's detection based on row and column differences catches.

\begin{tcolorbox}[title=\faIcon{lightbulb}]
\fxb enables controlled experiments on context gathering for formula suggestions.
\end{tcolorbox}

\subsection{Representation Understanding}
\label{sec:understanding}

\begin{table*}[htb]
\centering
\caption{Content strategy baselines on \fxb.}
\label{tab:content-baselines-compact-sonnet-metrics}
\scriptsize
\begin{tabular}{clrrrrrrrrr}
\toprule
\textbf{Class} & \textbf{Representation} & \multicolumn{3}{c}{\textbf{GPT 5.2}} & \multicolumn{3}{c}{\textbf{Sonnet 4.6}} & \multicolumn{3}{c}{\textbf{Qwen3-8B}} \\
\cmidrule(lr){3-5} \cmidrule(lr){6-8} \cmidrule(lr){9-11}
 & & \accuracy & \accuracyUpper & \rangeOverlap & \accuracy & \accuracyUpper & \rangeOverlap & \accuracy & \accuracyUpper & \rangeOverlap \\
\midrule
\multirow{5}{*}{\centering F}
& HTML                      & 58.3 & 58.3 & \textbf{67.6} & 60.0 & 60.0 & \textbf{71.0} & \textbf{20.7} & \textbf{20.7} & \textbf{36.1} \\
& CSV                       & 49.7 & 49.7 & 63.5 & 53.3 & 53.3 & 64.9 & 9.3 & 9.3 & 21.6 \\
& JSON                      & 56.9 & 56.9 & 67.1 & \textbf{60.4} & \textbf{60.4} & 69.7 & 14.5 & 14.5 & 30.6 \\
& Markdown                  & 53.3 & 53.3 & 66.1 & 54.7 & 54.7 & 67.5 & 11.3 & 11.3 & 25.2 \\
& Pipe-delimited & \textbf{58.8} & \textbf{58.8} & 66.3 & 59.8 & 59.8 & 69.9 & 12.1 & 12.1 & 27.4 \\
\midrule
\multirow{2}{*}{\centering W}
& Viewport                  & \textbf{56.9} & 82.4 & \textbf{64.9} & \textbf{58.4} & 84.7 & 67.4 & 14.1 & 20.5 & \textbf{27.9} \\
& SpreadsheetCoder          & 52.9 & \textbf{89.6} & 62.7 & 57.1 & \textbf{96.7} & \textbf{68.0} & \textbf{14.7} & \textbf{24.9} & 27.3 \\
\midrule
\multirow{2}{*}{\centering D}
& SheetCompressor           & \textbf{54.7} & \textbf{57.8} & \textbf{64.8} & \textbf{50.7} & 53.6 & \textbf{63.7} & 11.3 & 12.0 & 25.4 \\
& MiniSheet                 & 42.3 & 48.0 & 54.3 & 49.7 & \textbf{56.3} & 62.8 & \textbf{12.5} & \textbf{14.2} & \textbf{26.5} \\
\bottomrule
\end{tabular}
\par\smallskip
\end{table*}


To evaluate context (and representation) understanding, we perform zero-shot experiments on all formats from Section~\ref{ssec:context}, with five strategies for representing the full sheet: HTML, CSV, JSON, Markdown and pipe-delimited.
Table~\ref{tab:content-baselines-compact-sonnet-metrics} shows results for GPT 5.2 (non-reasoning) and Claude Sonnet 4.6 as representative frontier models, and Qwen3-8B as a small open model.
The relevant metrics are accuracy relative to upper bound (\accuracyUpper, to separate gathering from understanding) and the \% of references that was correctly recovered (\rangeOverlap, can the model infer data references from context).

\paragraph{Dense representations expose an understanding bottleneck.}

Minisheet, despite having among the highest theoretical coverage ($+$29.3\% over SpreadsheetCoder and $+$19.3\% over Viewport), is much less usable by frontier models.
SpreadsheetCoder improves over Minisheet by $+$41.6\% \accuracyUpper for GPT 5.2 and $+$40.4\% for Sonnet 4.6.
Windowed methods are mostly limited by missing evidence, while dense methods are limited by whether the model can understand the representation.

\paragraph{Mentioned ranges are not yet reliably usable.}

Minisheet's compression comes from mentioning references instead of printing them: at $w=2$, reference mention coverage is $+$61.2\% higher than reference print coverage.
However, these mentioned ranges are not reliably grounded back into cell addresses.
For GPT 5.2, Minisheet has $-$10.6\% \rangeOverlap compared to Viewport and $-$13.3\% compared to HTML; Sonnet 4.6 shows the same trend ($-$4.6\% and $-$8.2\%).
The main bottleneck is therefore dense numeric range clusters not being understood.
Minisheet is a good representation to study if we can teach these representations (through fine-tuning or prompting).

\paragraph{Models are (still) sensitive to representation.} 

Even when representing the exact same information, newer frontier models still suffer from divergent performance on different table representations \citep{singha2023tabular} with GPT 5.2 ($-$9.1\% \accuracy) and Sonnet 4.6 ($-$7.1\% \accuracy) having significant differences between the best and worst format---they also do not agree.
Minisheet uses a pipe-delimited representation, which is a compact mapping-based format that works well (JSON is mapping based).

\begin{tcolorbox}[title=\faIcon{lightbulb}]
\fxb enables research on (1) finding representations that are dense and understandable, and (2) training models to understand dense representations.
\end{tcolorbox}

\subsection{Understanding intent}
\label{sec:oracle}

To analyze the intent understanding capability, we perform zero-shot inference with oracle context.
The \textbf{base} oracle context contains the referenced cells and their values, the function names, and the annotated intent cells.
Seven variants are (1) + cell format (if available), (2) + dependency chain (of formula references and intermediate values), (3) + same column (other formulas in target column with their computed values), (4) + neighbor (5 $\times$ 7 viewport), (5) + all formulas in the worksheet, (6) + all of the above, (7) + an LLM annotated natural language description of the target formula.
The results are in Table~\ref{tab:oracle-main}. Further details, examples and qualitative analysis is discussed in Appendix~\ref{app:oracle-analysis}.

\begin{figure}[tb]
\centering
\begin{minipage}[t]{0.40\textwidth}
\centering
\captionof{table}{Oracle results on \fxb (GPT 5.2). Variants are additive over base. TP does not include abstention.}
\label{tab:oracle-main}
\scriptsize
\begin{tabular}{lrrr}
\toprule
\textbf{Oracle variant} & \textbf{TP} & \textbf{\accuracy} & $\boldsymbol{\Delta}$ \\
\midrule
base                 & 57.1 & 66.2 & --- \\
\midrule
\quad + cell format    & 57.1 & 66.2 & $+$0.0 \\
\quad + dependency chain      & 63.2 & 72.4 & $+$6.2 \\
\quad + same column    & 63.4 & 70.8 & $+$4.6 \\
\quad + neighbor       & 67.4 & 70.6 & $+$4.4 \\
\quad + formulas       & 66.2 & 71.4 & $+$5.2 \\
\midrule
all of the above       & 72.4 & 76.9 & $+$10.7 \\
\midrule
\quad + NL             & 88.1 & 97.2 & $+$31.0 \\
\bottomrule
\end{tabular}
\end{minipage}\hfill
\begin{minipage}[t]{0.59\textwidth}
\centering
\captionof{figure}{Failure modes }
\includegraphics[width=\textwidth]{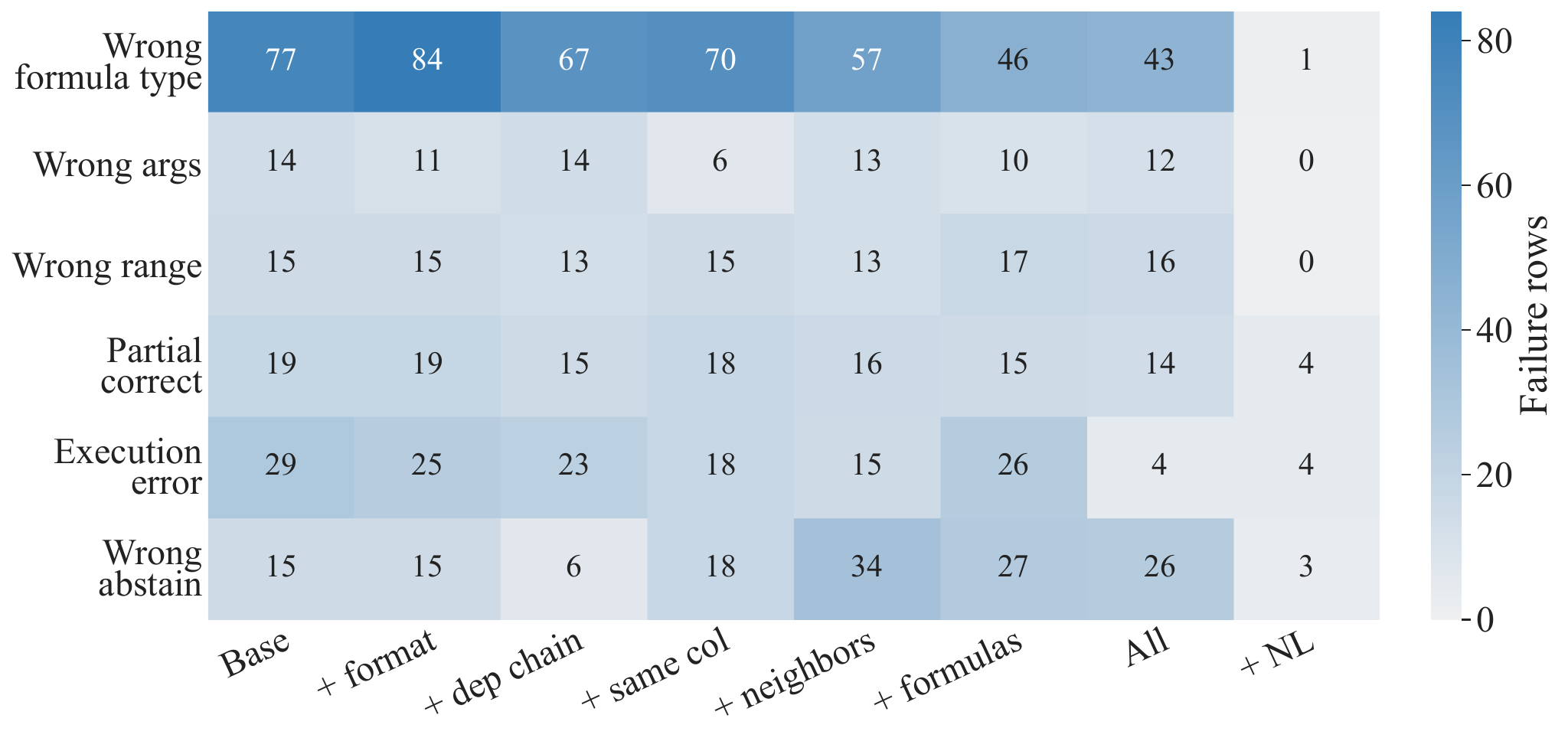}
\label{fig:oracle-analysis}
\end{minipage}
\end{figure}

\paragraph{Formula writing is mostly intent understanding.}

With oracle evidence but no explicit description, GPT 5.2 reaches only 66.2\% \accuracy, despite being given the referenced cells, their values, function names, and intent cells.
This shows that the bottleneck is not simply producing valid spreadsheet syntax, but inferring what computation the author intended from ambiguous spreadsheet evidence.
The strongest structured signals are those that indirectly reveal intent: dependency chains ($+$6.2\%), worksheet formulas ($+$5.2\%), same-column formulas ($+$4.6\%), and neighbors ($+$4.4\%).
Formatting gives no gain, suggesting that surface-level presentation is much less useful than cues about how nearby values are computed.

\paragraph{Structured context provides weak intent supervision}

Spreadsheet context helps most when it reveals the intended operation: wrong formula type failures drop from 77 to 43 with all structured signals.
However, wrong ranges remain flat (15 to 16), partial matches only slightly improve (19 to 14), and wrong abstentions increase (15 to 26).
Thus, structured context often makes the computation more plausible, but not precise enough to determine exact arguments, ranges, or answerability.

\paragraph{Explicit intent nearly solves formula writing}

Adding natural language raises accuracy to 97.2\% ($+31.0\%$ over base), and collapses almost all failure modes: wrong formula type drops from 77 to 1, while wrong arguments and wrong ranges disappear.
This suggests that GPT 5.2 can usually write the formula once intent is explicit.
The remaining bottleneck is therefore intent understanding: turning labels, layout, values, and formulas into the intended operation. 

\begin{tcolorbox}[title=\faIcon{lightbulb}]
\fxb enables controlled experiments on intent understanding with oracle context, separating failures caused by missing evidence from failures caused by composing the wrong formula.
\end{tcolorbox}

\section{Fine-Tuning for Compact Context}
\label{sec:fine-tuning}

\begin{table}[b]
\centering
\caption{Combined fine-tuning results on \fxb, comparing model size on Curated training data and dataset choice at fixed Qwen3-8B scale using eight
  sampled completions per problem.}
\label{tab:ft-combined-results}
\scriptsize
\begin{tabular}{lllrrrr}
\toprule
\textbf{Comparison} & \textbf{Model} & \textbf{Training data} & \textbf{Records} & \textbf{pass@1} & \textbf{pass@3} & \textbf{best@8} \\
\midrule
Base context & Qwen3-4B & -- & -- & 10.8\% & 12.6\% & 13.9\% \\
Base context & Qwen3-8B & -- & -- & 9.7\% & 12.2\% & 14.5\% \\
Base context & Qwen3-14B & -- & -- & 10.0\% & 13.4\% & 15.5\% \\
\midrule
Model size & Qwen3-4B & Curated & 25{,}614 & 16.5\% & 25.7\% & 32.8\% \\
Model size / anchor & Qwen3-8B & Curated & 25{,}614 & 21.2\% & 30.4\% & 38.0\% \\
Model size & Qwen3-14B & Curated & 25{,}614 & 19.5\% & 30.3\% & 40.2\% \\
\midrule
Dataset & Qwen3-8B & Random & 25,000 & 11.3\% & 17.3\% & 22.1\% \\
Dataset & Qwen3-8B & Simple Only & 25{,}000 & 14.1\% & 21.2\% & 27.4\% \\
Dataset & Qwen3-8B & Complex Only & 25{,}000 & 14.8\% & 23.5\% & 31.2\% \\
Dataset & Qwen3-8B & Scaled  & 39{,}161 & 20.6\% & 30.5\% & 38.8\% \\
Dataset & Qwen3-8B & Scaled 2$\times$ & 50{,}000 & 19.6\% & 29.2\% & 37.4\% \\
\midrule
Closed-model reference & GPT-5.2 & -- & -- & 40.2\% & 48.0\% & 53.9\% \\
\bottomrule
\end{tabular}
\end{table}

To overcome the gap between Minisheet's high theoretical coverage and poor zero-shot usability, models must learn how to interpret dense compact context.
We therefore fine-tune Qwen3 \citep{qwen3} models directly on Minisheet and evaluate (1) if supervised training can turn compact open models into useful formula completion systems, (2) the effect of model size and dataset scaling.
The training data is constructed from the remaining Sheetpedia workbooks, excluding all workbooks used in manual review (8,277 candidates).
Appendix~\ref{app:finetuning} gives dataset construction and training details.
Table~\ref{tab:ft-combined-results} shows pass@1 and pass@3 (both over 8 generations) and best@8 (any correct in eight candidates correct).

\paragraph{Fine-tuning teaches representation and intent understanding.}

Zero-shot Qwen3 models remain near 10$\%$ pass@1 under Minisheet, suggesting that larger base models (1) do not automatically know how to use dense spreadsheet context, (2) struggle to infer intent from that context.
After supervised fine-tuning, however, all models improve substantially in pass@1 ($+$11.1\% for 8B) and show even higher gains when pass@3 ($+$18.2\% for 8B) and best@8 ($+$23.5\%) control for clarity about the intent.
This is not just the model learning to write harder formulas: in best@8 we see $+$12.9\% from simple formulas (details in Appendix~\ref{app:training-data-variants}) and $+$23.5\% from additional curation, indicating that the model is also learning the representation.

\paragraph{Performance begins to saturate within Sheetpedia.}

Increasing scale within the same data source yields only modest gains.
Scaling up curated data slightly improves over the base 25K dataset on sampled metrics, reaching the best pass@3 (30.5\%) and best@8 (38.8\%), but is $-$0.6\% lower on pass@1.
Scaling up even more (2$\times$) reduces all three metrics relative to Scaled ($-$1.0\% pass@1, $-$1.3\% pass@3, $-$1.4\% best@8).
Model scale shows a similar pattern: under the same Curated data, 14B is essentially on par with 8B, with $-$1.7\% pass@1, $-$0.1\% pass@3 and only $+$2.2\% best@8.
Together, these results suggest that progress is beginning to saturate on the remaining Sheetpedia distribution.
Further gains may require curating more diverse supervision from additional sources.

\paragraph{A remaining bottleneck is ranking intent, not generating candidates.}

Even after fine-tuning, pass@higher $k$ remains much stronger than pass@1.
For 8B on Curated, best@8 exceeds pass@1 by $+$16.8\%; for 14B, the gap is $+$20.7\%.
The model thus often produces a correct formula somewhere in a small set of samples.
That pattern is consistent with intent uncertainty: the model can often guess plausible computations, but cannot reliably identify which one the spreadsheet author intended.
Interestingly, fine-tuning does \emph{not} hurt diversity.
This motivates further work on generate-and-rank \citep{ni2023lever}, or top-$k$ suggestion UX for formula suggestions \citep{chen2025need}.

\begin{tcolorbox}[title=\faIcon{lightbulb}]
\fxb enables research on (1) teaching dense spreadsheet representations to (smaller) models for efficient formula suggestions, (2) synthetic data generation for formula suggestions, (3) intent disambiguation and ranking, (4) and user experience research on formula suggestions.
\end{tcolorbox}

\section{Limitations and Future Work}

\fxb has a few limitations, most notably, as (1) only considering suggestions within one worksheet, (2) the intent annotations do not consider partial intent sufficiency, and (3) being scoped to English workbooks only.
None of these limitations diminish our contributions: \fxb is a solid starting point to encourage research fast, cheap and qualitative formula suggestions.
Besides improving the benchmark, \fxb enables new research on different aspects and capabilities for in-line formula suggestions (marked with \faIcon{lightbulb}).

\section{Conclusion}

We introduce \fxb, a benchmark of 503 in-line formula suggestion problems that consist of (workbook, target cell, expected formula or abstention marker) triples.
Each problem is further annotated with (1) cells that describe the intent, (2) cells that describe data sources, (3) and six other pieces of oracle information; to facilitate controlled experiments on different capabilities: context gathering, representation understanding, intent understanding, and formula writing.
We perform experiments to motivate further research, by (1) highlighting that dense representations are less understandable, (2) that intent understanding and disambiguation is a key problem, (3) that we can teach models to understand dense representations and detecting candidate intents (more diversity after fine-tuning).

\bibliographystyle{plainnat}
\bibliography{references}

\newpage
\appendix{}
\section*{Appendices}

\section{\fxb}
\label{app:fxbench}

\subsection{\fxb Construction Details}

\subsubsection{Phase 1: Programmatic Filtering}

Beyond the file and sheet thresholds reported in \S\ref{main:fxbench-construction}, we also drop workbooks containing VBA macros, cap worksheet count per workbook at 50, and exclude sheets composed entirely of formulas with no input data or just formulas and no data.

 \S\ref{main:fxbench-construction} introduces sketch and signature de-duplication; we expand on
 the example side here.
 
 \textbf{Sketch examples.}
 \begin{center}\small
 \begin{tabular}{@{}r@{\hspace{1em}}c@{\hspace{1em}}l@{}}
 \texttt{=SUM(A1:A10)} & $\rightarrow$ & \texttt{SUM(<range>)} \\
 \texttt{=VLOOKUP(B1, C1:D20, 2, FALSE)} & $\rightarrow$ & \texttt{VLOOKUP(<cell>, <range>, <int>, <bool>)} \\
 \texttt{=IF(A1>0, A1*0.1, 0)} & $\rightarrow$ & \texttt{IF(<cell> > <int>, <cell> * <float>, <int>)} \\
 \end{tabular}
 \end{center}
 
 \textbf{Signature example.}
 \begin{center}\small
 \texttt{=IF(SUM(A:A)>0, VLOOKUP(B1, C:D, 2, FALSE), 0)}
 \;$\rightarrow$\;
 \texttt{(IF, SUM, VLOOKUP) | cmp | 2}
 \end{center}
 Three functions in nesting order, one comparison-operator group,
 nesting depth two.
 
 \textbf{Hierarchy.} Counts at each abstraction level:
 \begin{center}\small
 \begin{tabular}{@{}lcccc@{}}
 & functions & signatures & sketches & formulas \\
 \midrule
 count & 148 & 15{,}845 & 136K & 24.6M \\
 \end{tabular}
 \end{center}
 
 \textbf{Why both levels.} Sketch de-duplication alone leaves many
 structurally identical patterns; signature de-duplication alone
 collapses useful argument diversity. The three formulas
 \begin{center}\small
 \texttt{=SUM(A1:A10)} \quad
 \texttt{=SUM(5, 10, 15)} \quad
 \texttt{=SUM(B1, C1, D1)}
 \end{center}
 all share signature \texttt{(SUM) |\;| 1} (one function, no
 operators, depth one) but their sketches---\texttt{SUM(<range>)},
 \texttt{SUM(<int>, <int>, <int>)}, and
 \texttt{SUM(<cell>, <cell>, <cell>)}---capture distinct authoring
 styles: a contiguous range, literal integers, and an explicit cell
 list. Both levels are needed to control redundancy while preserving
 this kind of variety.

\subsubsection{Phase 2: Annotation and Ranking}
\label{app:benchmark-phase-2}

\textbf{Programmatic annotation} computes structural attributes derivable directly from the workbook.

\begin{figure}[h]
\centering
\includegraphics[width=\linewidth]{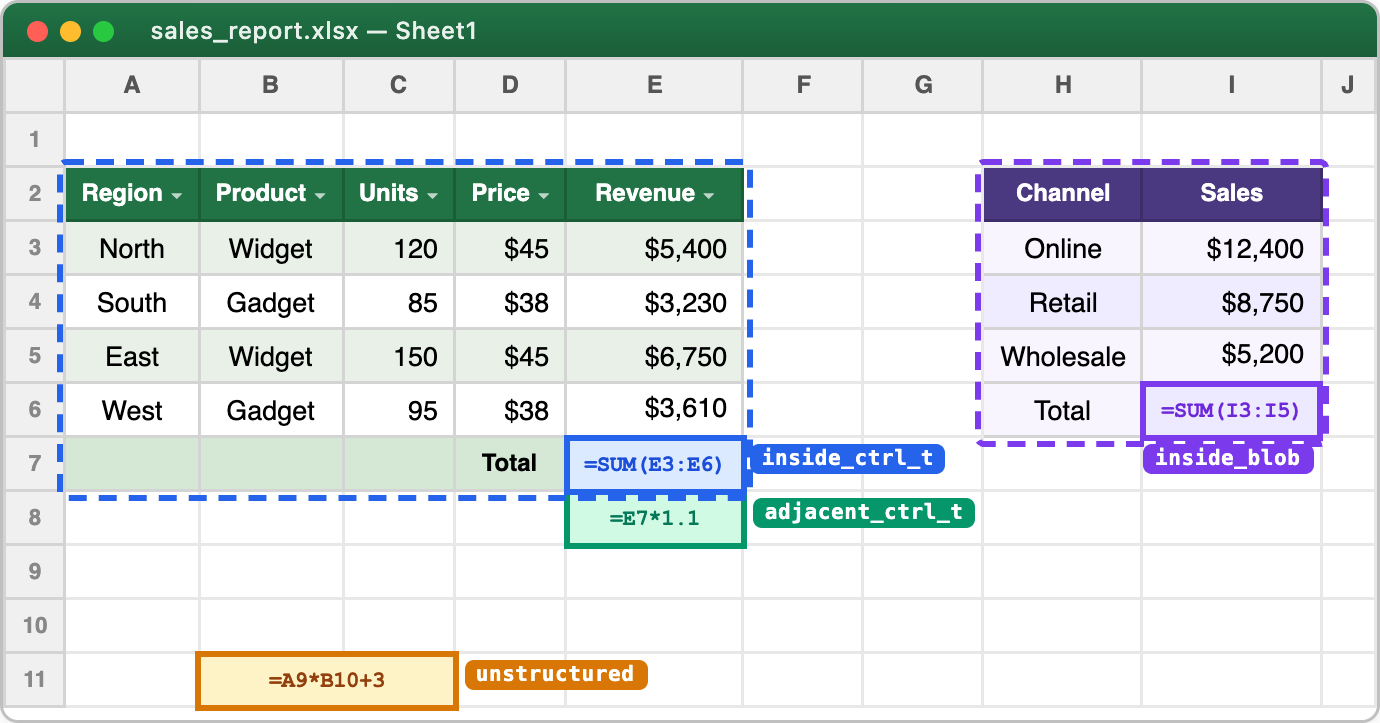}
\caption{Target cell location categories on a single worksheet.
 Dashed blue marks a structured Excel Table (Ctrl+T); dashed purple
 marks a blob. Each formula is color-coded by its location class:
 \texttt{=SUM(E3:E6)} is \texttt{inside\_ctrl\_t},
 \texttt{=SUM(I3:I5)} is \texttt{inside\_blob},
 \texttt{=E7*1.1} is \texttt{adjacent\_ctrl\_t}, and
 \texttt{=A9*B10+3} is \texttt{unstructured}.}
 \label{fig:cell-location}
 \end{figure}

\emph{Target cell location.} Each target cell is classified by its position relative to structured regions (Figure \ref{fig:cell-location}):  \texttt{inside\_ctrl\_t} (inside a structured Excel Table), \texttt{inside\_blob} (inside a contiguous rectangular region that is not a formal Table), \texttt{adjacent\_ctrl\_t} / \texttt{adjacent\_blob} (immediately bordering such a region but not part of it), and \texttt{unstructured} (neither inside nor adjacent to any structured region).

 \emph{Reference types.} Each formula is annotated with the types
 of references it uses (not mutually exclusive):
 \texttt{cell\_ref} (\texttt{A1}),
 \texttt{range\_ref} (\texttt{A1:A10}),
 \texttt{absolute\_ref} (\texttt{\$A\$1}),
 \texttt{cross\_sheet\_ref} (\texttt{Sheet2!A1}),
 \texttt{whole\_row\_col\_ref} (\texttt{A:A}), and
 \texttt{named\_ref} (\texttt{revenue\_total}).
 
\emph{Sheet-size bucket.} The worksheet is assigned to a size
 bucket based on total cell count, enabling downstream analysis of
 model performance across sheet scales.

\textbf{LLM-based intent cue classification} annotates each record with its strongest available intent hint. For each record, we extract a $\pm 15$ row, $\pm 10$ column window around the target cell, serialize it as a markdown table with the target cell marked as \texttt{[TARGET]} and formulas preserved as strings, and pass it to GPT-5.2 using the prompt in Listing \ref{fig:intent-cue-prompt}. The judge returns three fields: location classification, intent hint source, and a one-sentence description of what the formula computes. These labels describe what cues are present, and these are only used as an axis for sampling to curate a balanced distribution across intent cue types. Table \ref{tab:distributions} reports the distribution across \fxb. 

\begin{listing}
\caption{Prompt used for LLM-based intent cue classification. Template variables \texttt{\{markdown\_table\}}, \texttt{\{cell\_location\}}, and \texttt{\{formula\}} are filled per record.}
 \label{fig:intent-cue-prompt}
\begin{gblocksmall}
You are an expert spreadsheet analyst. You will be given a spreadsheet region 
as a markdown table, a target cell location, and the formula in that cell.
 
The target cell is marked with [TARGET] in the table. Formulas in other cells are 
shown as their formula strings (e.g., =B2+C2), not their computed values.
 
## Spreadsheet Context
{markdown_table}
 
**Target cell:** {cell_location}
**Formula:** {formula}
 
## Annotations Required
 
### 1. Location Classification
- `table_cell` (*@\textemdash@*) Inside a structured table or rectangular data region with clear headers
- `table_adjacent` (*@\textemdash@*) Within 5 cells of a structured table
- `unstructured` (*@\textemdash@*) No obvious tabular structure nearby
 
### 2. Intent Hint
Pick the single strongest hint source:
- `col_header` (*@\textemdash@*) A column header describes the intent
- `row_label` (*@\textemdash@*) A row label describes the intent
- `adjacent_formula` (*@\textemdash@*) Neighboring cells contain similar formulas suggesting a pattern
- `adjacent_values` (*@\textemdash@*) A pattern in adjacent cell values hints at the computation
- `combination` (*@\textemdash@*) Multiple hint sources contribute equally
- `no_hint` (*@\textemdash@*) No clear contextual clue available
 
### 3. Intent Description
In one short sentence, describe what this formula computes.
 
Respond ONLY with JSON:
{
    "location": "...", 
    "intent_hint": {"source": "...", "hint_text": "..."},
    "intent_description": "..."
}
\end{gblocksmall}
\end{listing}

\textbf{Intent-aware de-duplication} groups record by \texttt{(signature, intent hint source)}, and we caps formulas per group. Informative hint categories \texttt{(col\_header, row\_label, adjacent\_formula, combination)} are capped at 3 records per signatures, weaker hint categories \texttt{(adjacent\_values, no\_hint)} at 5, preserving coverage of structurally unusual records. Within each group, records are drawn from distinct workbooks via shuffled round-robin sampling with a fixed seed.

\textbf{Greedy diversity selection}\label{app:greedy-selection} operates within each primary-function bucket, defined by the outermost function in the formula (e.g., \texttt{SUMIFS}, \texttt{VLOOKUP}, \texttt{IF}), capped
 at 100 records per function. At each step, the candidate maximizing a composite score is selected and the score components are updated before the next pick. The score combines four signals:
 \begin{itemize}
     \item \textbf{Signature rarity}: rewards previously unseen structural signatures.
     \item \textbf{Intent-source rarity}: favors under-represented hint categories.
     \item \textbf{Workbook novelty}: prefers records from workbooks not yet selected.
     \item \textbf{Nesting complexity}: a small bonus for deeper formulas.
 \end{itemize}
 Functions like \texttt{IF} and bare arithmetic expressions exhibit wide internal variety (e.g., simple conditionals vs.\ deeply nested ones), so we subdivide their budgets proportionally across
 complexity tiers before applying greedy selection.


\subsubsection{Phase 3: Manual Curation and Verification}
\label{app:temporal}

The third phase converts ranked candidates into a verified benchmark through three steps.

\begin{figure}[h]
    \centering
    \includegraphics[width=\linewidth]{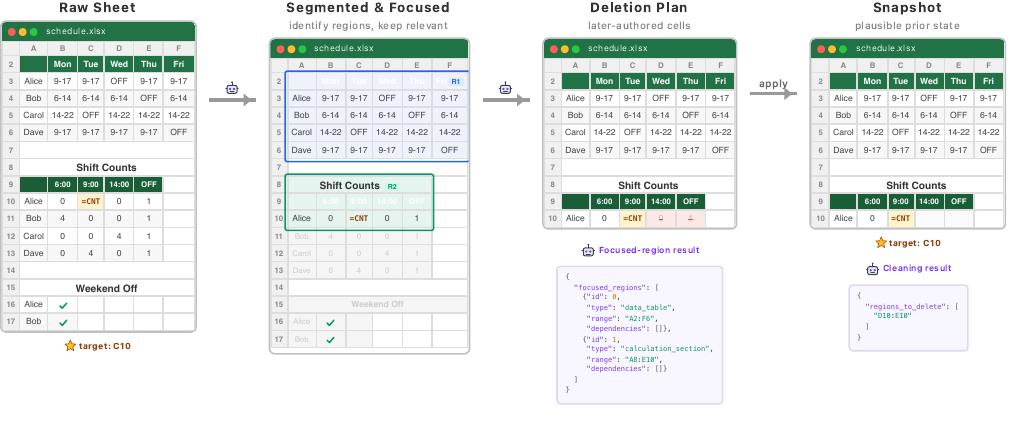}
    \caption{Temporal snapshot pipeline}
    \label{fig:snapshot-pipeline}
\end{figure}

\textbf{Temporal Snapshots.} A spreadsheet at corpus collection time reflects its final state, including formulas authored after the target cell. Cells later in the sheet may follow obvious patterns, allowing a model to copy rather than predict. No prior spreadsheet benchmark controls for this. We generate a plausible historical snapshot of each workbook; an approximation of the state the user likely saw before entering the target formula. We use a two pass LLM pipeline illustrated in Figure \ref{fig:snapshot-pipeline}.

\paragraph{Pass 1: Segmentation.} The target sheet is rendered as a dense markdown table with Excel-style row and column labels. An LLM is asked to partition all non-empty cells into non-overlapping logical regions, each with a type, range and dependency list (Figure \ref{fig:snapshot-prompts}). The pipeline then identifies the region containing the target cell and reconstructs a focused version of the sheet, retaining only the regions the formula depends on and all data tables. The result is a chronologically ordered subset of the sheet containing only the context that would have existed at the time the user was about to write the formula. 

\paragraph{Pass 2: Cleaning.} The focused workbook is re-rendered to dense markdown and a second LLM identifies ranges to delete (Figure \ref{fig:snapshot-prompts}). The cleaning pass assumes a plausible authoring order: top-to-bottom, left-to-right. It deletes cells sharing the same logical role as the target that appear later in authoring order, while preserving cells referenced by the formula, headers and labels, raw input regions, and cells serving different logical roles. When uncertain, the LLM is instructed to prefer preservation over deletion. Deleted cells are blanked without shifting coordinates, keeping all formula references stable.

\begin{figure}[p]
     \centering
     \includegraphics[width=\linewidth]{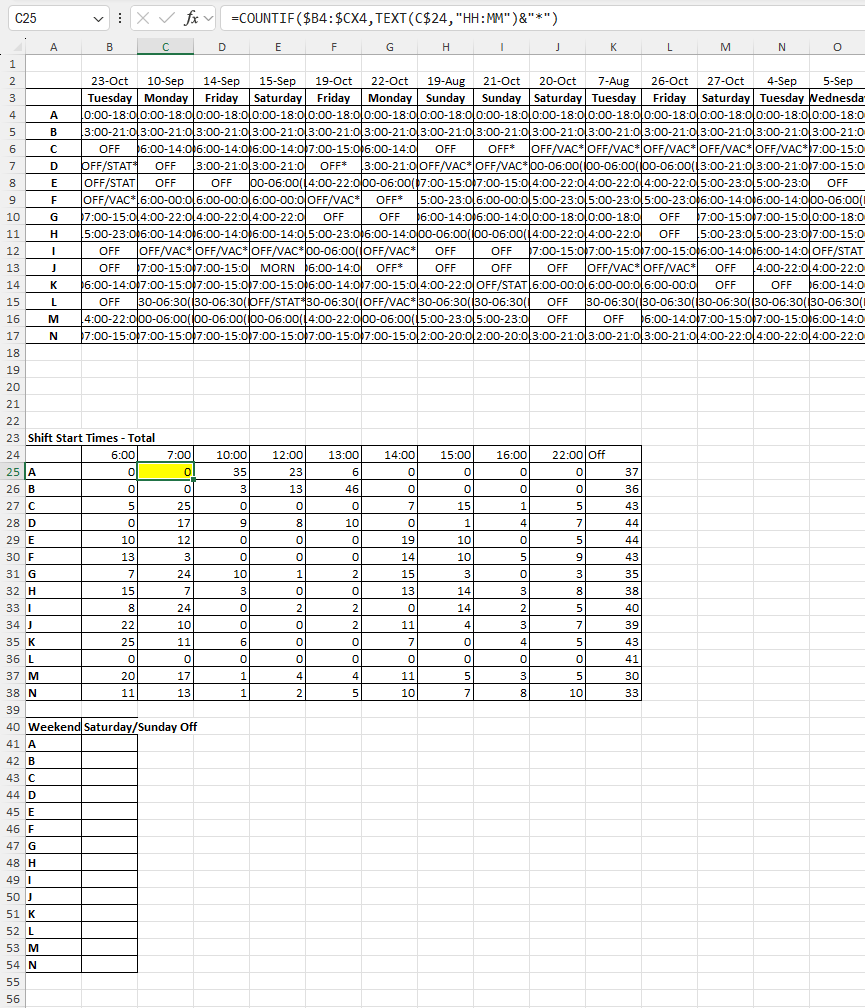}
     \caption{Temporal snapshot: before.}
     \label{fig:snapshot-before}
 \end{figure}

\begin{figure}[p]
     \centering
     \includegraphics[width=\linewidth]{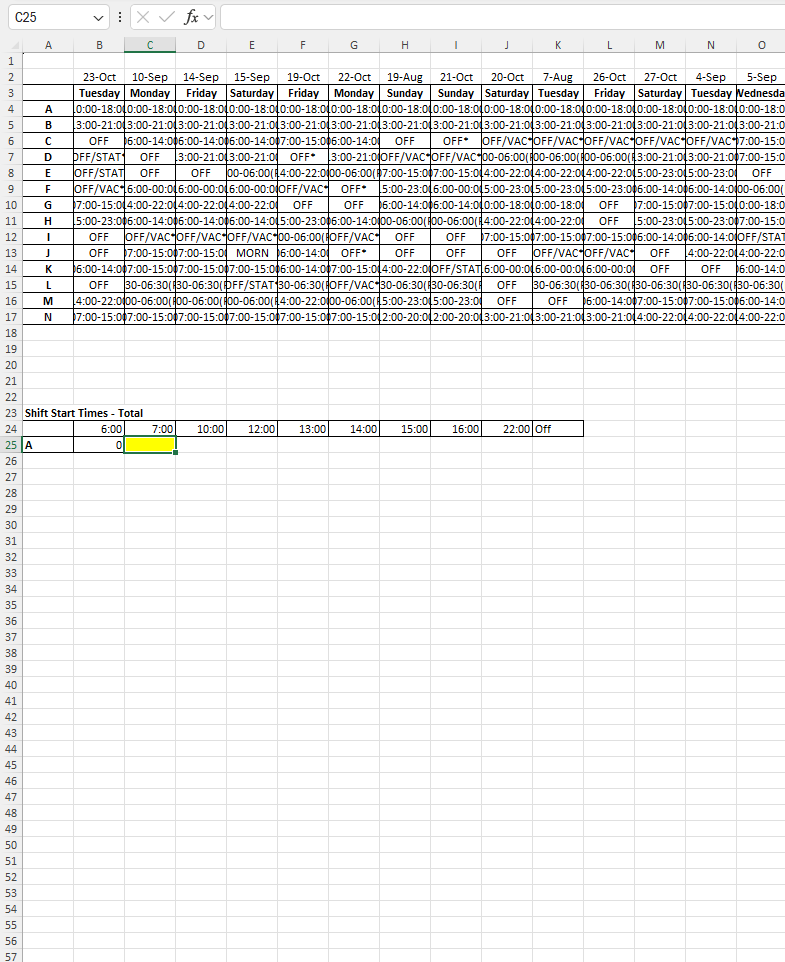}
     \caption{Temporal snapshot: after. }
     \label{fig:snapshot-after}
 \end{figure}

Figure \ref{fig:snapshot-before} and Figure \ref{fig:snapshot-after} shows a real before-and-after 
example from FxBench.

\begin{figure*}
\begin{minipage}[t]{0.48\linewidth}
\begin{gblocksmall}
You are an expert Excel analyst. Identify logical segments in the sheet.
 
Region Types:
- "data_table": Structured table with headers
- "calculation_section": Formulas/computed results separate from main tables
- "sheet_title_header": Main title/heading
- "text_description": Notes, instructions
- "miscellaneous": Other content
 
 Principles:
 1. Separate different business purposes
 2. Follow content to its natural end
 3. Use section headers for groupings
 4. Prioritize logical function over layout
 
 Requirements:
 - Unique integer 'id' starting from 0
 - 'range' in Excel notation (e.g. "A1:G32")
 - Cover ALL non-empty cells, no overlap
 - Include 'dependencies' listing region IDs
 
 Output:
 { "regions": [
   {"id":0, "range":"A1:G32",
    "type":"data_table", "dependencies":[]},
   {"id":1, "range":"J7:K9",
    "type":"calculation_section",
    "dependencies":[0]}
]}
 \end{gblocksmall}
 \centering\scriptsize (a) Segmentation (Pass 1).
 \end{minipage}
 \hfill
 \begin{minipage}[t]{0.48\linewidth}
 \begin{gblocksmall}
You are an expert Excel analyst. Identify cell ranges 
to DELETE to create a historical snapshot--the state 
BEFORE a specific formula was written.
 
Core Assumptions:
1. Users populate top-to-bottom, left-to-right
2. Raw data entered BEFORE referencing formulas
3. Earlier time periods before later ones
 
Deletion Rules:
Delete cells logically written after the
active cell. Use headers/labels to identify
the active cell's role (e.g. "Total", "Sum").
Delete cells sharing that role authored after.
 
Authoring order:
- Reading: same row right, then rows below
- Temporal: later period columns come later
 
Preservation Rules (Never Delete):
- Cells referenced by the active formula
- Row/column headers providing context
- Data regions the formula operates on
- Raw input data regions
- Cells with different logical roles
 
Output:
{
    "regions_to_delete":["A10:D20","G15"]
}
 
NEVER delete referenced cells. When uncertain, prefer preservation.
 \end{gblocksmall}
 \centering\scriptsize (b) Cleaning (Pass 2).
 \end{minipage}
 \caption{System prompts for the temporal snapshot pipeline.}
 \label{fig:snapshot-prompts}
 \end{figure*}

\textbf{Saturation pruning}. Beyond the four criteria of manual review listed in \S\ref{main:fxbench-construction}, we also drop records whose pattern is already sufficiently represented---e.g., column totals such as \texttt{=SUM(B2:B10)} that recur in near-identical form across many workbooks.

\textbf{ABSTAIN records}. A small subset of candidates are flagged as ABSTAIN, target cells where no formula should reasonably be written or where intent is ambiguous even to a human reviewer. For example, a cell adjacent to a data table with no header, no row label, and no siblings formulas to suggest what computation in expected.  These test a model's ability to recognize when context is insufficient and withhold a prediction than guess.

\subsection{Release Statement}

\fxb will be released under \texttt{CC BY-SA 4.0}, inherited from Sheetpedia. We respect the original license and redistribute content only within the terms they allow. The construction pipeline and evaluation code will also be publicly released.

 \begin{figure}
     \centering
     \includegraphics[width=0.55\textwidth]{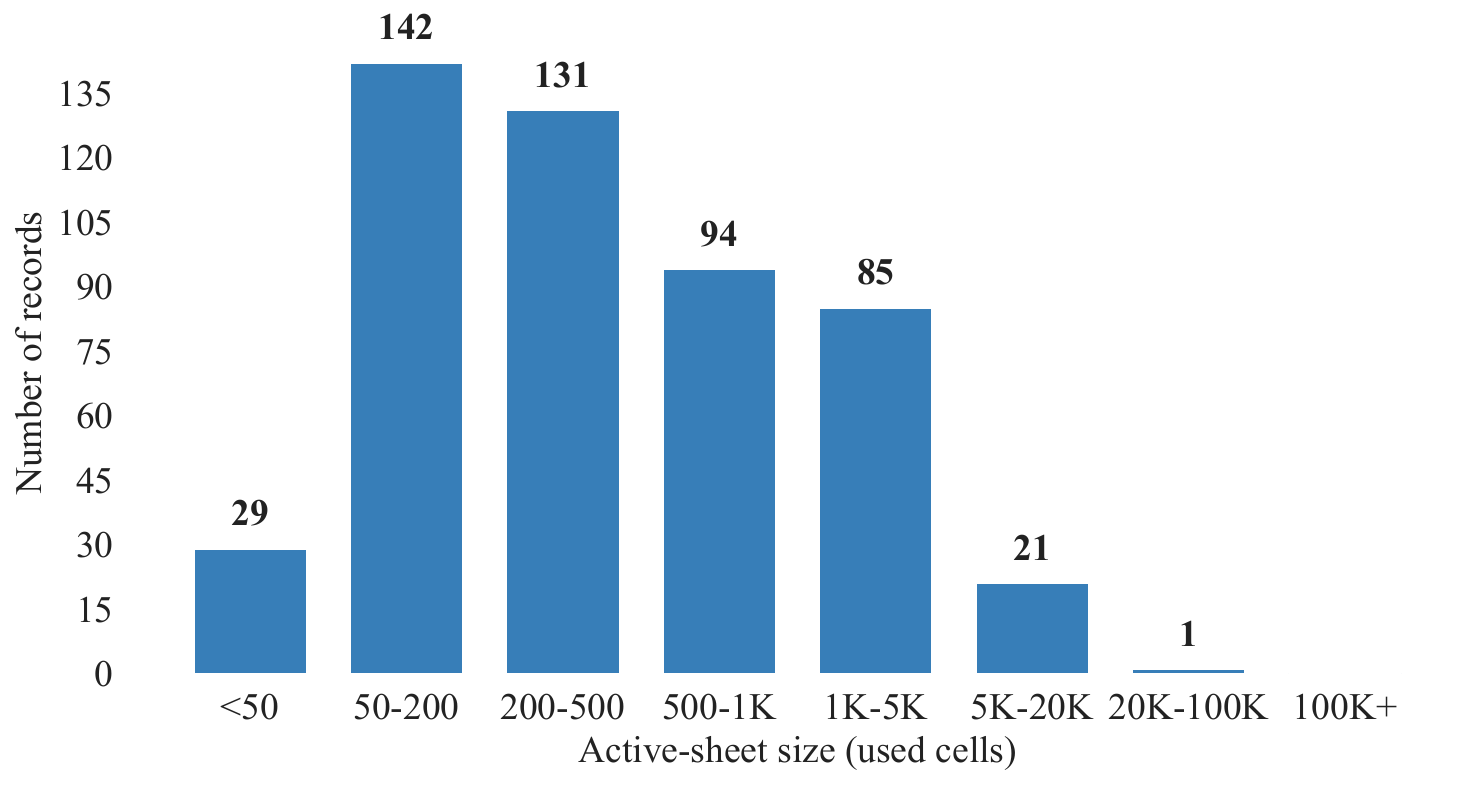}
     \caption{Sheet Size Distribution }
     \label{fig:sheet-size}
 \end{figure}

\begin{table}[t]
\centering
\small
\caption{Exhaustive per-category function inventory of \fxb. Every distinct function is listed under its Excel category with its record count in parentheses.}
\label{tab:function-inventory}
\begin{tabular}{@{}llll@{}}
\toprule
\multicolumn{4}{@{}l@{}}{\textbf{Statistical} \quad (36 functions, 139 records)} \\
AVERAGE (17) & COUNTIFS (16) & COUNTIF (15) & MIN (13) \\
MAX (12) & LARGE (10) & COUNTA (7) & COUNT (4) \\
INTERCEPT (4) & SMALL (4) & AVERAGEIF (3) & AVERAGEIFS (3) \\
FORECAST (3) & MINIFS (3) & LINEST (2) & STDEV.P (2) \\
MAXA (2) & AVERAGEA (1) & CORREL (1) & COUNTBLANK (1) \\
FORECAST.LINEAR (1) & MAXIFS (1) & MINA (1) & MODE.MULT (1) \\
NORM.S.INV (1) & PEARSON (1) & PERCENTILE.INC (1) & QUARTILE.EXC (1) \\
RSQ (1) & STDEV.S (1) & TRIMMEAN (1) & MEDIAN (1) \\
SLOPE (1) & TREND (1) & VAR.S (1) & STDEVA (1) \\
\midrule
\multicolumn{4}{@{}l@{}}{\textbf{Math \& trigonometry} \quad (38 functions, 128 records)} \\
ABS (22) & SUMPRODUCT (14) & SUMIF (13) & SUM (8) \\
AGGREGATE (6) & INT (6) & EXP (5) & MOD (5) \\
ROUNDUP (4) & MROUND (3) & CEILING (2) & CEILING.MATH (2) \\
COS (2) & FLOOR (2) & LN (2) & MMULT (2) \\
ROUNDDOWN (2) & SQRT (2) & SUBTOTAL (2) & SUMIFS (2) \\
POWER (2) & QUOTIENT (2) & LOG (2) & SERIESSUM (2) \\
ACOS (1) & ATAN (1) & DEGREES (1) & GCD (1) \\
LOG10 (1) & PRODUCT (1) & ROUND (1) & SIN (1) \\
SUMSQ (1) & SUMXMY2 (1) & TRUNC (1) & SIGN (1) \\
TAN (1) & RADIANS (1) &  &  \\
\midrule
\multicolumn{4}{@{}l@{}}{\textbf{Date \& time} \quad (20 functions, 49 records)} \\
HOUR (9) & DATE (6) & DATEDIF (6) & DAY (4) \\
DATEVALUE (3) & MONTH (3) & DAYS (2) & EDATE (2) \\
NETWORKDAYS (2) & TIMEVALUE (2) & DAYS360 (1) & MINUTE (1) \\
NETWORKDAYS.INTL (1) & YEAR (1) & WORKDAY (1) & EOMONTH (1) \\
WORKDAY.INTL (1) & WEEKNUM (1) & YEARFRAC (1) & WEEKDAY (1) \\
\midrule
\multicolumn{4}{@{}l@{}}{\textbf{Lookup \& reference} \quad (9 functions, 43 records)} \\
LOOKUP (11) & INDEX (9) & MATCH (7) & VLOOKUP (6) \\
HLOOKUP (4) & XLOOKUP (3) & CHOOSE (1) & INDIRECT (1) \\
TRANSPOSE (1) &  &  &  \\
\midrule
\multicolumn{4}{@{}l@{}}{\textbf{Financial} \quad (19 functions, 33 records)} \\
FV (6) & NPV (5) & IRR (3) & PRICE (2) \\
PV (2) & DB (2) & ACCRINT (1) & COUPNCD (1) \\
COUPPCD (1) & CUMIPMT (1) & FVSCHEDULE (1) & IPMT (1) \\
NPER (1) & PMT (1) & PPMT (1) & RATE (1) \\
SLN (1) & YIELD (1) & XIRR (1) &  \\
\midrule
\multicolumn{4}{@{}l@{}}{\textbf{Logical} \quad (5 functions, 28 records)} \\
IF (11) & IFERROR (7) & AND (5) & IFS (3) \\
IFNA (2) &  &  &  \\
\midrule
\multicolumn{4}{@{}l@{}}{\textbf{Text} \quad (7 functions, 11 records)} \\
CONCAT (3) & RIGHT (2) & TRIM (2) & CODE (1) \\
CONCATENATE (1) & VALUE (1) & REPT (1) &  \\
\midrule
\multicolumn{4}{@{}l@{}}{\textbf{Compatibility} \quad (7 functions, 9 records)} \\
QUARTILE (3) & CONFIDENCE (1) & MODE (1) & PERCENTILE (1) \\
RANK (1) & STDEV (1) & STDEVP (1) &  \\
\midrule
\multicolumn{4}{@{}l@{}}{\textbf{Database} \quad (2 functions, 2 records)} \\
DMAX (1) & DSUM (1) &  &  \\
\midrule
\multicolumn{4}{@{}l@{}}{\textbf{Engineering} \quad (1 functions, 1 records)} \\
CONVERT (1) &  &  &  \\
\midrule
\multicolumn{4}{@{}l@{}}{\textbf{Information} \quad (1 functions, 1 records)} \\
N (1) &  &  &  \\
\midrule
\multicolumn{4}{@{}l@{}}{\textbf{Abstain} \quad (1 functions, 49 records)} \\
ABSTAIN (49) \emph{(no formula expected)} &  &  &  \\
\midrule
\multicolumn{4}{@{}l@{}}{\textbf{Expression} \quad (1 functions, 10 records)} \\
EXPRESSION (10) \emph{(no top-level function)} &  &  &  \\
\bottomrule
\end{tabular}
\end{table}

\clearpage
\section{Context Representations}  

\subsection{Coverage Analysis}
\label{app:theoretical-analysis}

We define a family of coverage metrics that measure how completely a context strategy exposes the cells a formula depends on. For each \fxb problem $i$ we compare \emph{required} cell sets against what the strategy renders, and report the fraction of problems whose required cells are all exposed.

\paragraph{Required sets.}
\begin{itemize}
    \item $D_i$: \emph{direct\_intent} cells---headers and labels that hint at what the formula computes.
    \item $A_i$: \emph{data\_description} cells describing the input data being used.
    \item $I_i = D_i \cup A_i$: all intents cells for problem $i$.
    \item $R_i$: cells referenced by the ground-truth formula.
\end{itemize}
We drop empty cells and the target cell itself from every required set, since every strategy blanks the target by design. The intent sets come from the \fxb annotations \S\ref{app:intent-annotation}.

\paragraph{Exposed sets.} For strategy $s$ on problem $i$ we track two cell sets:
\begin{itemize}
    \item $P_{i,s}$ (\emph{printed}):cells appearing on their own with their exact address and exact value or formula text. Compressed ranges and type-aggregated tuples do not count as printed.
    \item $M_{i,s}$ (\emph{mentioned}): cells named or recoverable from the output---an explicit address, a range or cluster region containing the cell, or a header notation pointing at its location. By construction $P_{i,s} \subset M_{i,s}$.
\end{itemize}

\paragraph{Coverage metrics.} Let $N$ be the number of benchmark problems (503). We do not give partial credit: one missing cell fails the check. The five reported metrics are:
 \begin{align*}
 \textsc{IntentPrinted}(s)    &= \tfrac{1}{N}\sum_{i=1}^{N}
   \mathbf{1}\!\left[\, I_i \subseteq P_{i,s} \,\right] \\
 \textsc{IntentMentioned}(s)  &= \tfrac{1}{N}\sum_{i=1}^{N}
   \mathbf{1}\!\left[\, I_i \subseteq M_{i,s} \,\right] \\
 \textsc{RefPrinted}(s)       &= \tfrac{1}{N}\sum_{i=1}^{N}
   \mathbf{1}\!\left[\, R_i \subseteq P_{i,s} \,\right] \\
 \textsc{RefMentioned}(s)     &= \tfrac{1}{N}\sum_{i=1}^{N}
   \mathbf{1}\!\left[\, R_i \subseteq M_{i,s} \,\right] \\
 \textsc{Upper}(s)            &= \tfrac{1}{N}\sum_{i=1}^{N}
   \mathbf{1}\!\left[\, D_i \subseteq P_{i,s}
         \,\land\, (A_i \cup R_i) \subseteq M_{i,s} \,\right]
 \end{align*}

 The first four ask whether each required set ($I_i$ or $R_i$) is fully \emph{printed} or merely \emph{mentioned}. \textsc{Upper} captures the realistic ceiling. Table \ref{tab:context-evidence-metrics-short-v4} reports these metrics together with token cost on \fxb.

\subsection{Intent Cells Annotation}
\label{app:intent-annotation}

The intent cells used by the structural coverage metrics are the cells a model must wire together to arrive at the formula. They are the headers and row labels needed either to identify parts of the spreadsheets serve as inputs to the formula or to describe what the formula is supposed to compute. 

\textbf{Annotation procedure.} For each problem we build a small local context for the annotator by collecting all non-empty cells within $\pm4$ rows and columns of (1) the target cell, (2) each individual cell referenced by the ground-truth formula, and (3) each corner cells of every range referenced by the formula. Sampling around range corners, rather than every cell in the range. For a rectangular range such as \texttt{B6:F23} all four corners (\texttt{B6}, \texttt{F6}, \texttt{B23}, \texttt{F23}) contribute to their own local windows. Cell values are truncated to 32 characters. The annotator is prompted with the target cell, the ground-truth formula, and this local context as show in Figure \ref{fig:intent-prompt}.


\textbf{Why an LLM annotator.} While \texttt{data\_range} is directly derivable from the ground-truth formula, identifying intent headers and row labels for each formula requires manual analysis. Most real-world spreadsheets do not use structured tables, data is messy and unstructured, so simple positional heuristics are unreliable, and doing this by hand across the benchmark is time-consuming. 

\begin{listing}
\caption{Prompt used for intent cell annotation. Template
 variables \texttt{\{target\_cell\}}, \texttt{\{formula\}},
 \texttt{\{context\}}, and \texttt{\{data\_ranges\}} are
 filled per FxBench record. \texttt{\{context\}} is a
 newline-separated list of \texttt{Address~|~value} pairs
 for non-empty cells in local windows around the target
 cell and around the corners of each formula-referenced
 range. \texttt{\{data\_ranges\}} is a comma-separated list
 of formula ranges, or \texttt{None (individual cell
 references only)} when the formula contains no ranges.}
\label{fig:intent-prompt}
\begin{gblocksmall}
 [System]
 You are an expert at identifying cells in Excel that help understand formula intent.
 
 Given a formula, its target cell, and surrounding cell context, identify  three types of cells and the formula intent:
 
 1. Direct Intent: Cells describing what the OUTPUT/RESULT represents
    - Column header above the target cell (where the formula lives)
    - Row labels next to the target cell
    - Focus: What is being CALCULATED or PRODUCED
    - Example: If target is G6 with header "Percent Rank" at G2, return ["G2"]
 
 2. Data Description: Cells describing the INPUT data being used
    - Column headers for ranges/cells referenced IN the formula
    - Row labels for data being consumed
    - Focus: What data is being READ or CONSUMED
    - Example: For =F6/SUM(F6:F23), if F2 is "Percent Male", return ["F2"]
 
 3. Data Range: Full range notations used in the formula
    - Use colon notation: "F6:F23" not separate boundaries
    - Include all ranges found in the formula
    - Example: For SUM(F6:F23), return ["F6:F23"]
 
 4. Intent: A concise description (at most 10 words) of
    what the formula calculates.
    - Example: "Total sales for east region"
 
 Return cell addresses as a JSON object with four fields:
 {
    "direct_intent": ["G2"],
    "data_description": ["F2"],
    "data_range": ["F6:F23"],
    "intent": "Percentage of male respondents in the survey"
 }
 
 Important guidelines:
 - Return actual cell addresses, not descriptions
 - Direct intent describes the target cell's meaning (OUTPUT)
 - Data description describes the referenced data's meaning (INPUT)
 - Data range uses colon notation for ranges
 - Use the context to identify relevant headers and labels
 - Return all categories; use empty lists if nothing applies
 
 [User]
 # Target Cell
 {target_cell}
 
 # Formula
 {formula}
 
 # Context
 The following cells surround the formula target and referenced cells:
 {context}
 
 # Data Ranges
 The formula operates on these ranges: {data_ranges}
\end{gblocksmall}
\end{listing}

\subsection{Oracle Analysis}
\label{app:oracle-analysis}


\subsubsection{Oracle Variants}

All oracle variants share a common \textbf{Base} context: the gold formula's referenced cells with their values, the formula's function names, the annotated intent cells, and the target marked as \texttt{[TARGET]}. Cell values are grouped by exact value or by structural pattern, and formula cells show their computed values. Base removes the comprehension load of locating the relevant cells while preserving the model's task of assembling them into a formula.

We define seven variants over Base:

\textbf{+ cell format:} the target cell's expected display format (percentage, currency, date, time, etc) when detectable.

\textbf{+ dependency chain:} the recursive dependency chain of the formula's references, including intermediate computed values.

\textbf{+ same column:} other formulas in the target's column with their computed values, capturing drag-down patterns.

\textbf{+ neighbor:} a $5\times7$ grid centered on the target cell, capturing local spatial layout.

\textbf{+ formulas:} every formula in the worksheet, grouped by structural pattern, exposing sheet-wide conventions.

\textbf{+ all of the above:} stacks all of the above, the strongest oracle using only spreadsheet derived signals.

\textbf{+ NL Prompt:} a natural-language description of the target formula and how to wire its arguments. Prompts are LLM-generated; we treat this as an upper bound where the model is told what to compute. Refer \ref{app:nl-prompt-generation} for NL prompt generation details.

\subsubsection{Oracle Variant Examples}
\label{app:oracle-prompts}

We illustrate the oracle variants on a single \fxb problem: the
 target cell \texttt{Data!E26} on a workbook tracking starting-amps
 measurements by side and status. The expected formula is
 {\small\verb|=AVERAGE(AVERAGEIF(E3:E23,{"*side a*","*low*"}, I3:I23))|}.
 Figure~\ref{fig:oracle-prompts} shows \textbf{Base} in full,
 followed by the additions made by \textbf{all of the above} and
 \textbf{+NL Prompt}. Workbook contents are PII-redacted; status
 strings shown as \texttt{HIGH\_COLLECTING\_DATA} and
 \texttt{LOW\_COLLECTING\_DATA} are placeholder tokens.

 \begin{figure}
 \textbf{\small Base Oracle Prompt (full)}
 \begin{gblocksmall}
 Sheet: Data
 Target cell: E26
 Intent: Average starting amps for side A or low status
 Consider using: AVERAGE
 
 Intent cells:
   E2,I2     | Status, Starting Amps Value      | input_header
   E25,D26   | SKU, Collected Value Average     | output_header
 
 Referenced cells (grouped by value):
   E3,E6:E7,E9,E11,E13,E16:E17      | HIGH_COLLECTING_DATA
   E4:E5,E8,E10,E12,E14:E15,E18,E20 | LOW_COLLECTING_DATA
   E19,E21   | side b, side a
   I3..I21   | numeric values (16 entries)
 
 E26 | [TARGET]
 Data ranges: E3:E23, I3:I23
 \end{gblocksmall}
 \vspace{0pt}
 \textbf{\small + all of the above (additions over Base)}
 \begin{gblocksmall}
 Consider using: AVERAGE, AVERAGEIF        [function hint expanded]
 
 Computation chain:
   E3 = HIGH_COLLECTING_DATA;  I3 = 1.5831
   E23 = (empty);              I23 = (empty)
   -> E26 = [TARGET]
 
 All sheet formulas (grouped by pattern):
   F26 | =AVERAGE(AVERAGEIF(E3:E23,{"*side a*","*low*"},  M3:M23))
   G26 | =AVERAGE(AVERAGEIF(E3:E23,{"*side a*","*low*"},  P3:P23))
   J26 | =AVERAGE(AVERAGEIF(E3:E23,{"*side b*","*high*"}, I3:I23))
   K26,L26 | (same pattern over M3:M23, P3:P23)
   ...   (15 unrelated MAXIFS/MINIFS/COUNT patterns omitted)
 \end{gblocksmall}
 \textbf{\small +NLPrompt (additions over Base)}
 \begin{gblocksmall}
 Formula instruction: In E26, return the average starting amps for rows where the status in E3:E23 contains either "side a" or "low" anywhere in the text. Use AVERAGEIF with an array of those two wildcard criteria against E3:E23 and average the matching values from I3:I23, then wrap that result in AVERAGE so the two criteria results are combined into one final average.
 \end{gblocksmall}
 
 \caption{Oracle variants for the same \fxb problem
 (\texttt{Data!E26}). The first panel shows the full Base prompt;
 the \emph{+ all of the above} and \emph{+NL Prompt} panels show only the lines they
 add. }
 \label{fig:oracle-prompts}
 \end{figure}

\subsubsection{NL Prompts for Formulas}
\label{app:nl-prompt-generation}


For each FxBench record, an annotator LLM (GPT-5.2) receives the ground-truth formula, target cell, and intent annotation and a small surrounding context window. The window is built from a 3 cell radius around the target and each cell referenced in the formula. Each context cell is serialized as \texttt{Address | value} with values truncated to 32 characters. The annotator's system prompt is shown in Listing \ref{fig:nl-annotator-prompt}

\begin{listing}
\begin{gblocksmall}
 [System]
 You are an expert Excel formula instructor. Given a target cell, the ground-truth  Excel formula,  and surrounding worksheet context, write a concise natural-language instruction that another Excel expert could use to reconstruct the formula exactly.
 
 Your instruction must include:
   1. Goal     - what the target cell should display.
   2. Method   - which Excel functions to use and how the logic connects.
   3. Data     - exact cells/ranges to operate on.
   4. Tricky   - exact string literals, argument order, omitted arguments, array behavior, where multipliers go, and any non-obvious wiring.
 
 Constraints:
   - Do not just emit the formula text.
   - Do not be vague about string literals, ranges, or argument order.
   - Keep the explanation to 1-4 sentences.
 
 Return JSON:
 {"nl_prompt": "...", "complexity": "simple|moderate|complex"}
 
 [User]
 # Formula
 {ground_truth_formula}
 
 # Target Cell
 {target_cell}
 
 # Intent
 {intent_text_if_available}
 
 # Surrounding Context
 {local_context}
\end{gblocksmall}
 \caption{System and user prompt template used to generate the
 natural-language instructions consumed by the \textsc{+NL Prompt}
 oracle.}
 \label{fig:nl-annotator-prompt}
\end{listing}

\begin{table}[t]
 \centering
 \caption{Deterministic failure labels used in the oracle failure analysis. Labels are assigned only to actual failures (FP/FN/Error); correct 
predictions and correct abstentions are excluded.}
 \label{tab:oracle-failure-labels}
 \scriptsize
 \setlength{\tabcolsep}{4pt}
 \begin{tabular}{p{0.31\linewidth}p{0.62\linewidth}}
 \toprule
 \textbf{Failure label} & \textbf{Description} \\
 \midrule
 \texttt{incorrect\_formula\_type}
 & The prediction uses a different root function or a substantially different nested-function strategy from the target. \\
 
 \texttt{right\_type\_incorrect\_args}
 & The root and nested function structure are broadly correct, but non-range arguments such as criteria, constants, operators, argument order, or
 wrapper arguments are wrong. \\
 
 \texttt{right\_type\_incorrect\_range}
 & The root and nested function structure are broadly correct, but one or more referenced cells or ranges are wrong. \\
 
 \texttt{partial\_correct}
 & The prediction shares the outer formula strategy and some nested structure, but an important nested subexpression is missing, added, or 
replaced. \\
 
 \texttt{abstain\_wrong}
 & The model makes an abstention mistake: it abstains when a formula is expected, or produces a formula when abstention is expected. \\
 
 \texttt{syntax\_error}
 & The predicted formula cannot be parsed or accepted as a valid Excel formula. \\
 
 \texttt{execution\_error}
 & The predicted formula is syntactically valid but evaluates to an Excel runtime error such as \texttt{\#N/A}, \texttt{\#VALUE!}, or 
\texttt{\#NUM!}. \\
 
 \texttt{other\_formula\_mismatch}
 & A valid, executable formula mismatch that does not fit the deterministic type, range, argument, or partial-structure rules. \\
 \bottomrule
 \end{tabular}
 \end{table}




\clearpage
\newpage
\subsubsection{Qualitative Analysis}
\label{app:qualitative}

\paragraph{Case 1: Missing the intent-hint}

\begin{multicols}{2}
 \small
 The intent says \emph{moving} average and a window length lives
 in \texttt{C2} (=15), but the model returns the raw aggregation
 \gcode{=MAX(G3:G310)} over the full column. It latches onto the
 function noun (\texttt{MAX}) and ignores the windowing modifier,
 defaulting to the simplest aggregation that matches the verb.

 The expected formula:
 \begin{gblocksmall}
 =MAX(SUBTOTAL(1,OFFSET(G2:G3,ROW(G3:G310)-ROW(G3),,C2)))
 \end{gblocksmall}
 constructs a trailing \texttt{C2}-row window at every offset via
 \texttt{OFFSET}+\texttt{SUBTOTAL} and takes the max over those
 window averages. The cue is local (\texttt{C2}=15 plus the
 \emph{Highest Moving Average} header), so this is a
 context-utilization failure rather than under-specification.
 
 \columnbreak
 
 \begin{gblocksmall}
 # Context
 
 Sheet: Max Moving Average (3)
 Target cell: B2
 Intent: Highest moving average for P
 Consider using: MAX
 
 Intent cells:
   B1 | Highest Moving Average | output_header
   G1 | P | input_header
   C1 | C | input_header
   A2 | Highest Moving Average P | output_header
 
 Referenced cells (grouped by value):
   C2,G145,G199 | 15
   G2:G21,G25:G26,G50,G53:G55,... | 0
   ... [truncated]
 
 B2 | [TARGET]
 
 Data ranges: G2:G3, G3:G310
 
 # Task
 
 Current cell: "B2"
 Prefix: `=`
 \end{gblocksmall}
 \end{multicols}
 
 \begin{figure}[h]
 \centering
 \includegraphics[width=\linewidth]{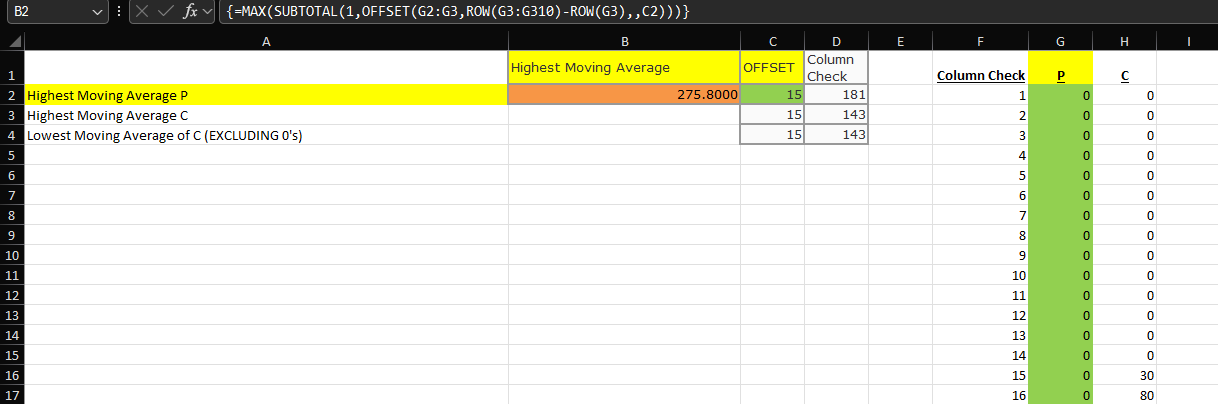}
 \caption{Sheet view for Case 1 (\texttt{Max Moving Average!B2}).
 The formula bar shows the trailing-window construction;
 \texttt{C2} holds the window length (15).}
 \label{fig:case1-moving-max}
 \end{figure}

 \paragraph{Case 2: Missing tie-breaking over matched candidates.}
 
 \begin{multicols}{2}
 \small
 The model recovers the broad lookup structure---find the minimum
 price in \texttt{B2:F2} and return the corresponding quantity from
 \texttt{B3:F3}---with
 
 \begin{gblocksmall}
 =INDEX(B3:F3,MATCH(MIN(B2:F2),B2:F2,0))
 \end{gblocksmall}
 
 \noindent However, the minimum price (50.00) appears for multiple
 suppliers (Sup2, Sup4). The predicted \texttt{MATCH} returns the
 first tied supplier (Sup2, quantity 50), while the expected formula
 
 \begin{gblocksmall}
 =LARGE(IF($B$2:$F$2=MIN($B$2:$F$2),$B$3:$F$3),1)
 \end{gblocksmall}
 
 \noindent considers all minimum-price suppliers and returns the
 \emph{largest} corresponding quantity (Sup4, quantity 70). The
 \textbf{Consider using: LARGE} hint in the prompt points directly
 at this construction; the model ignores it. This is a hint-utilization
 failure: the tie-breaking rule is not visually obvious from the data,
 but the function hint exposes it.
 
 \columnbreak
 
 \begin{gblocksmall}
 # Context
 
 Sheet: Sheet2
 Target cell: B6
 Intent: Quantity corresponding to the lowest supplier price
 Consider using: LARGE
 
 Intent cells:
   A2 | Price | input_header
   A3 | Quantity | input_header
   A6 | lowest price | output_header
 
 Referenced cells (grouped by value):
   B2,D2,F2 | 172.5
   C2:C3,E2 | 50
   B3 | 90
   D3 | 40
   E3 | 70
   ... [truncated]
 
 B6 | [TARGET]
 
 Data ranges: B2:F2, B3:F3
 
 # Task
 
 Current cell: "B6"
 Prefix: `=`
 \end{gblocksmall}
 \end{multicols}
 
 \begin{figure}[h]
 \centering
 \includegraphics[width=\linewidth]{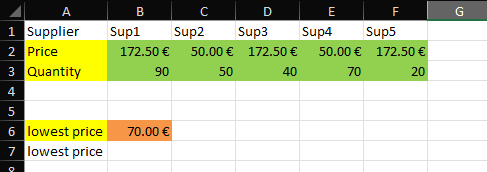}
 \caption{Sheet view for Case 2 (\texttt{Sheet2!B6}). The minimum price
 50.00 ties between Sup2 (quantity 50) and Sup4 (quantity 70); the
 expected formula breaks the tie by selecting the larger quantity.}
 \label{fig:case2-tiebreak}
 \end{figure}

 \paragraph{Case 3: Right formula, incorrect range.}
 
 The model selects the right statistical operation
 (\gcode{=VAR.S(...)}) but grounds it to a single visible column
 \gcode{E4:E13} instead of the full data block.
 The expected formula is \gcode{=VAR.S(B4:F13)}. The data range
 \texttt{B4:F13} is named verbatim in the prompt header
 (\textbf{Data ranges}), so this is a clean context-utilization
 failure: the model anchors on the column closest to the target
 cell rather than the explicitly enumerated range.
 
 \smallskip
 \noindent\begin{minipage}[t]{0.48\linewidth}
 \vspace{0pt}
 \begin{gblocksmall}
 # Context
 
 Sheet: Example 1.3-answers
 Target cell: E19
 Consider using: VAR.S
 
 Intent cells:
   D19 | Variance = | input_header
 
 Referenced cells (grouped by value):
   B4 | 990
   C4 | 4610
   D4 | 5740
   E4 | 7170
   F4 | 9010
   ... [truncated]
 
 E19 | [TARGET]
 
 Data ranges: B4:F13
 
 # Task
 
 Current cell: "E19"
 Prefix: `=`
 \end{gblocksmall}
 \end{minipage}%
 \hfill
 \begin{minipage}[t]{0.48\linewidth}
 \vspace{0pt}
 \centering
 \includegraphics[width=\linewidth]{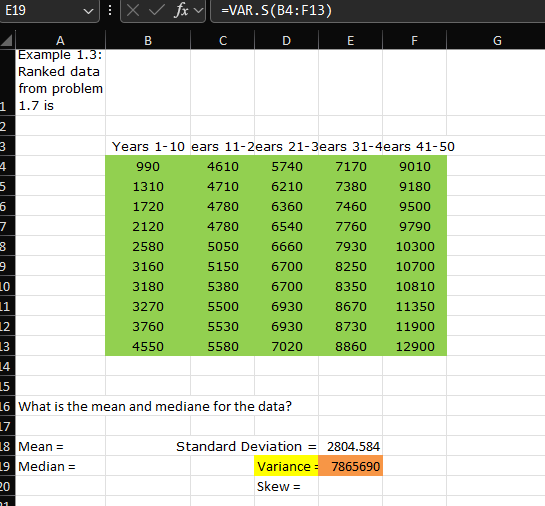}
 \captionof{figure}{Sheet view for Case 3}
 \label{fig:case3-wrong-range}
 \end{minipage}

\paragraph{Case 4: Right formula, wrong rate frequency.}
 
 \begin{figure}
 \centering
 \includegraphics[width=0.75\linewidth]{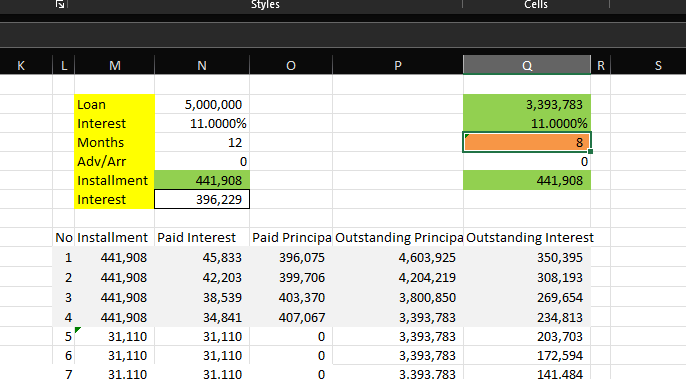}
 \caption{Sheet view for Case 4 (\texttt{restruct!Q4}). Interest is
 quoted as an annual rate (11.0000\%) in \texttt{Q3} but the output
 column is \emph{Months}, so \texttt{NPER} needs the monthly rate
 \texttt{Q3/12}.}
 \label{fig:case4-rate-freq}
 \end{figure}
 
 \begin{multicols}{2}
 \small
 The model recovers the right financial function and argument
 order with \gcode{=NPER(Q3,Q6,-Q2,0,Q5)}, while the expected
 formula is \gcode{ =NPER(Q3/12,Q6,-Q2,0,Q5)}
 
 \noindent The mismatch is unit reconciliation: \texttt{Q3} holds an
 annual interest rate (11\%), but the output header \emph{Months}
 implies the period count must be in months, so the rate must be
 divided by 12. The model picks up the rate cell by header match
 (\textbf{Interest}) but does not reconcile its unit against the
 output column. The model fails to operationalize the intent---the unit
  mismatch is invisible from the cell values alone and only the
  juxtaposition of the two header semantics
  (\emph{Interest}=annual, \emph{Months}=monthly periods) reveals it

 
 \columnbreak
 
 \begin{gblocksmall}
 # Context
 
 Sheet: restruct
 Target cell: Q4
 Intent: Number of months to repay the loan
 Consider using: NPER
 
 Intent cells:
 M2 | Loan | input_header
 M3 | Interest | input_header
 M4 | Months | output_header
 M5 | Adv/Arr | input_header
 M6 | Installment | input_header
 
 Referenced cells (grouped by value):
 
 Formulas (grouped by pattern):
 Q2 | =P13
 Q3,Q5:Q6 | =N@
 
 Q4 | [TARGET]
 
 # Task
 
 Current cell: "Q4"
 Prefix: `=`
 \end{gblocksmall}
 \end{multicols}

\paragraph{Case 5: Partially correct (moving average over the wrong window).}
 
 \begin{figure}[h]
 \centering
 \includegraphics[width=\linewidth]{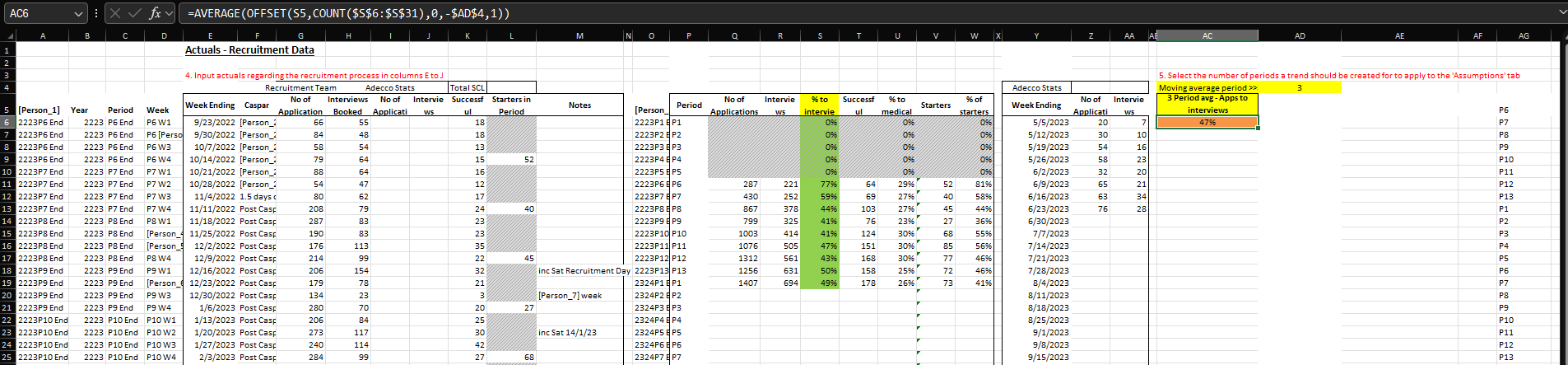}
 \caption{Sheet view for Case 5 (\texttt{Data\_Recruitment Act!AC6}).
 The expected formula bar shows a trailing window anchored to the last
 populated row of column \texttt{S}, with length \texttt{\$AD\$4}=3.}
 \label{fig:case5-partial}
 \end{figure}
 
 \begin{multicols}{2}
 \small
 The target asks for a 3-period moving average of \emph{\% to interview}
 and the model produces a structurally plausible prediction
 \gcode{=AVERAGE(OFFSET(S6,0,0,\$AD\$4,1))}: it correctly identifies the
 function (\texttt{AVERAGE}), the window-size parameter
 (\texttt{\$AD\$4}=3), and the relevant column (\texttt{S}). The expected
 formula is
 
 \begin{gblocksmall}
 =AVERAGE(OFFSET(S5,COUNT($S$6:$S$31),0,-$AD$4,1))
 \end{gblocksmall}
 
 \noindent which constructs a \emph{trailing} window: it anchors at the
  last populated cell of \texttt{S6:S31} via \texttt{COUNT} and walks
  backwards \texttt{\$AD\$4} rows using a negative height. The model's
  window starts at the current row and walks forward, so it averages
  future rather than past values---right shape, wrong direction.
  The intent line names a moving average over completed periods, but
  the model anchors at the current row and walks forward instead. We
  classify this as \emph{partially correct}.

 
 \columnbreak
 
 \begin{gblocksmall}
 # Context
 
 Sheet: Data_Recruitment Act
 Target cell: AC6
 Intent: Moving average of percentage to interview
 Consider using: AVERAGE
 
 Intent cells:
   AC4 | Moving average period >>> | input_header
   S5  | % to interview            | input_header
   AC5 | 3 Period avg - Apps to interv... | output_header
 
 Referenced cells (grouped by value):
   AD4 | 3
   S6:S10 | 0
 
 Formulas (grouped by pattern):
   S11:S31 | =IFERROR(R@/Q@,"")
 
 AC6 | [TARGET]
 
 Data ranges: S6:S31
 
 # Task
 
 Current cell: "AC6"
 Prefix: `=`
 \end{gblocksmall}
 \end{multicols}

\clearpage
\section{Fine-Tuning Details}
\label{app:finetuning}

\subsection{Training Data Variants}
\label{app:training-data-variants}

We construct fine-tuning datasets from Sheetpedia formulas after excluding, every workbook in Phase 3 manual review and every workbook appearing in the \fxb. From this remaining pool we build six training variants.

\textbf{Curated}. The Curated corpus is sampled by primary Excel function. We target up to 1000 formulas per primary function type. For small function pools, we keep all available records while limiting each workbook to at most five records per function type. For larger pools, we select 1000 records using a greedy diversity selection similar to Appendix \ref{app:greedy-selection}, prioritizing signature diversity, workbook diversity, and formula complexity. The final Curated corpus contains 25,614 records from 17,892 workbooks and 172 formula types.

\textbf{Simple-only}. Same target size as Curated, restricted to simple formulas: function depth $\leq 2$ length, $\leq 100$ commonly used functions, no cross-sheet references, signature frequency $\geq 10$ in the candidate pool. Final: 25,000 records from 13,533 workbooks across 59 formula types.

\textbf{Complex-only}. Counterpart to Simple-only: deeper nesting, longer formulas, rare function types, rare functions signatures, or cross-sheet references. Final: 25,000 records from 9,503 workbooks across 74 formula types.

\textbf{Curated-Scaled-Up}. Starts from Curated and adds 13,547 workbook-disjoint records via the same greedy procedure, raising the per-function cap to 2,000. Final: 39,161 records from 27,327 workbooks across 172 formula types.

\textbf{Curated-Scaled-2X}. Forced-scaling variant targeting 50K records: starts from Curated-Scaled-Up, then fills the remaining budget by sampling per primary function type. Larger but more concentrated among frequent types. Final: 50,000 records from 32,020 workbooks across 172 formula types.

\textbf{Random}. As an additional baseline, we create a dataset by random sampling records from the training pool without any function balancing or complexity targeting. 

Together, these variants let us study how fine-tuning is affected by three factors: formula diversity, formula complexity, and dataset scale.

\begin{figure}[b]
   \centering
   \includegraphics[width=0.48\linewidth]{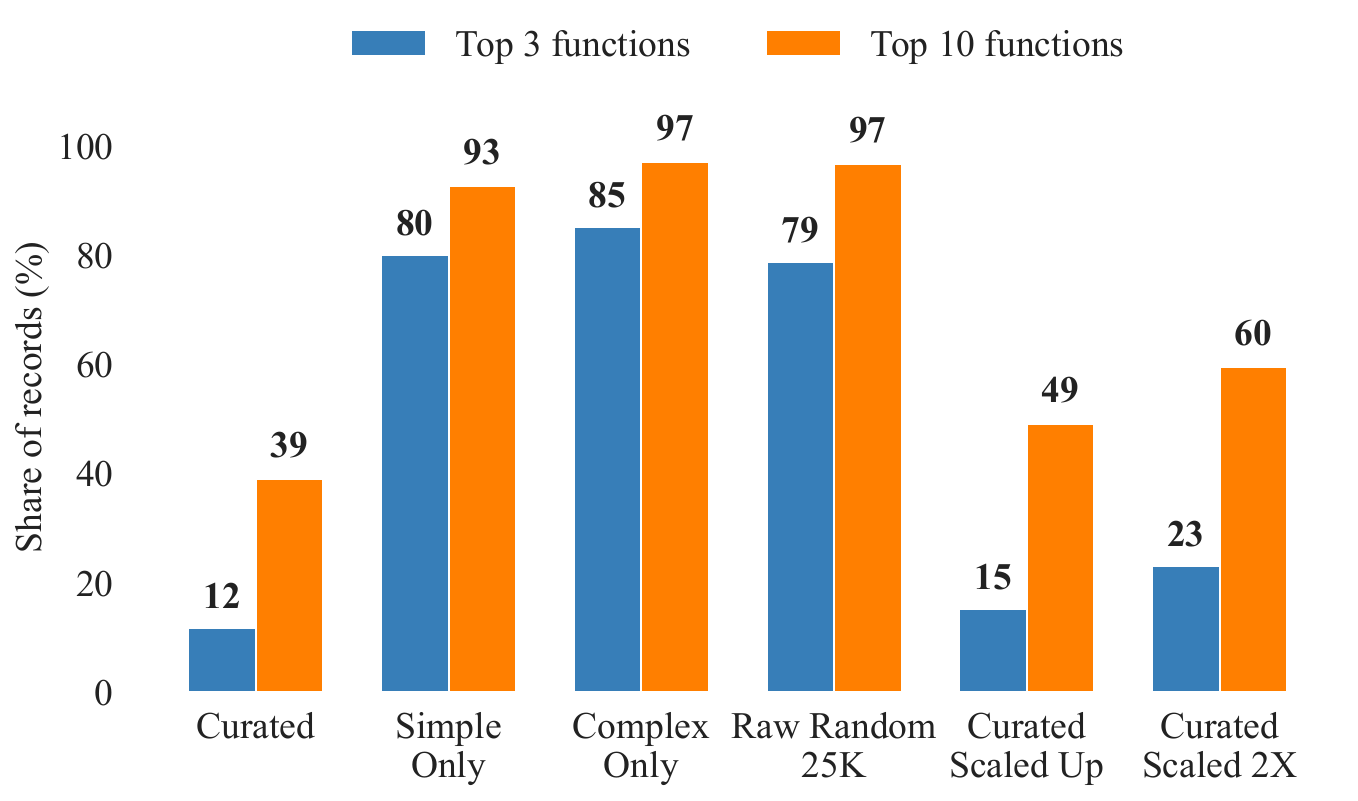}
   \includegraphics[width=0.48\linewidth]{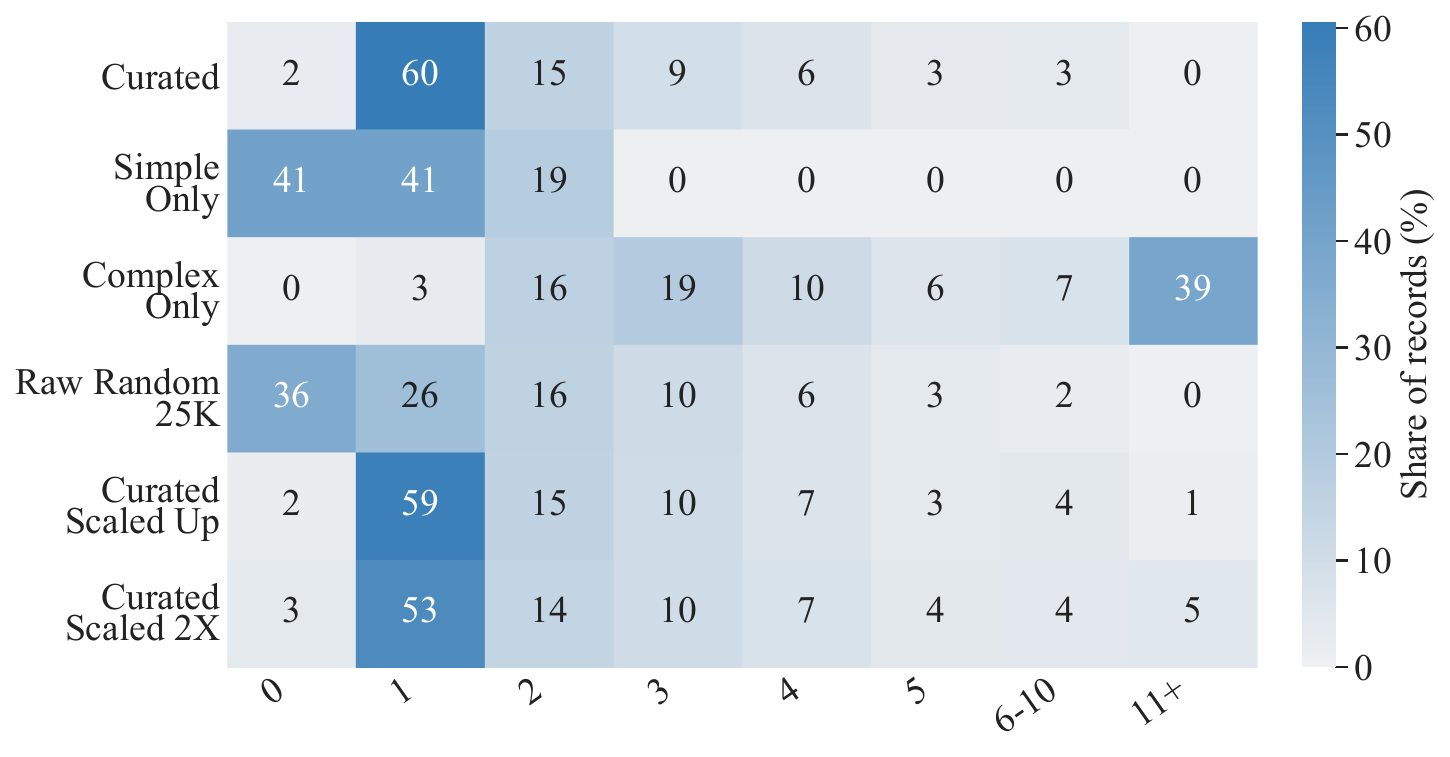}
   \caption{Left: share of records (\%) covered by the to three and top ten primary function types. Right: distribution of function nesting depth across dataset variants.}
   \label{fig:finetuning_dataset_stats}
 \end{figure}

 \subsection{Fine-Tuning Configuration}

 All fine-tuned Qwen3 models use full-parameter supervised fine-tuning on MiniSheet prompts. All runs use \texttt{AdamW 8-bit} optimization, peak learning rate $1\times10^{-5}$, cosine decay with 5\% warmup, weight decay 0.01, maximum gradient norm 1.0, maximum sequence length 4096, effective batch size 32, and unmasked prompt loss. For the Qwen3-8B data-variant experiments, Curated, Simple-only, and Complex-only are trained for 1100 steps; Curated-Scaled-Up is trained for 1800 steps and Curated-Scaled-2x is trained for 2200 steps. All fine-tuning experiments were run on 2 x H100 GPUs.

\clearpage
\begin{table*}[t]
\centering
\caption{Content strategy baselines on \fxb: outcome counts and auxiliary metrics across GPT-5.2, Sonnet-4-6, and Qwen3-8B.}
\label{tab:content-baselines}
\scriptsize
\setlength{\tabcolsep}{3pt}
\begin{tabular}{llrrrrrrrr}
\toprule
\textbf{Context Builder} & \textbf{Model} & \textbf{TP} & \textbf{FP} & \textbf{TN} & \textbf{FN} & \textbf{Acc\%} & \textbf{Overlap\%} & \textbf{Syntax\%} & \textbf{Sketch\%} \\
\midrule
\multicolumn{10}{l}{\textit{Full-sheet strategies}} \\
HTML                      & GPT-5.2 & 281 & 204 & 12 & 6 & 58.3 & \textbf{67.6} & 98.0 & \textbf{27.4} \\
                          & Sonnet-4-6 & 288 & 199 & 14 & 2 & 60.0 & \textbf{71.0} & 91.3 & \textbf{27.4} \\
                          & Qwen3-8B & 90 & 352 & 14 & 47 & \textbf{20.7} & \textbf{36.1} & 81.9 & \textbf{3.2} \\
\addlinespace[1pt]
CSV                       & GPT-5.2 & 236 & 243 & 14 & 10 & 49.7 & 63.5 & 96.8 & 25.6 \\
                          & Sonnet-4-6 & 251 & 230 & 17 & 5 & 53.3 & 64.9 & 91.8 & 24.9 \\
                          & Qwen3-8B & 38 & 396 & 9 & 60 & 9.3 & 21.6 & 78.3 & 2.0 \\
\addlinespace[1pt]
JSON                      & GPT-5.2 & 276 & 209 & 10 & 8 & 56.9 & 67.1 & \textbf{98.2} & 25.8 \\
                          & Sonnet-4-6 & 290 & 195 & 14 & 4 & \textbf{60.4} & 69.7 & 91.3 & 27.4 \\
                          & Qwen3-8B & 65 & 388 & 8 & 42 & 14.5 & 30.6 & 81.3 & 1.8 \\
\addlinespace[1pt]
Markdown                  & GPT-5.2 & 251 & 217 & 17 & 18 & 53.3 & 66.1 & 95.6 & 25.0 \\
                          & Sonnet-4-6 & 256 & 220 & 19 & 8 & 54.7 & 67.5 & 91.8 & 25.6 \\
                          & Qwen3-8B & 44 & 411 & 13 & 35 & 11.3 & 25.2 & \textbf{83.3} & 2.8 \\
\addlinespace[1pt]
Pipe delimited Full-sheet & GPT-5.2 & 285 & 199 & 11 & 8 & \textbf{58.8} & 66.3 & 98.0 & 25.8 \\
                          & Sonnet-4-6 & 284 & 199 & 17 & 3 & 59.8 & 69.9 & \textbf{92.2} & 26.6 \\
                          & Qwen3-8B & 50 & 395 & 11 & 47 & 12.1 & 27.4 & 81.5 & 2.6 \\
\midrule
\multicolumn{10}{l}{\textit{Windowed approaches}} \\
Viewport                  & GPT-5.2 & 274 & 212 & 12 & 5 & \textbf{56.9} & \textbf{64.9} & \textbf{98.6} & \textbf{26.8} \\
                          & Sonnet-4-6 & 275 & 202 & 19 & 7 & \textbf{58.4} & 67.4 & 91.1 & 27.4 \\
                          & Qwen3-8B & 61 & 410 & 10 & 22 & 14.1 & \textbf{27.9} & \textbf{86.1} & \textbf{2.2} \\
\addlinespace[1pt]
SpreadsheetCoder          & GPT-5.2 & 257 & 222 & 9 & 15 & 52.9 & 62.7 & 95.6 & 25.0 \\
                          & Sonnet-4-6 & 263 & 206 & 24 & 10 & 57.1 & \textbf{68.0} & \textbf{92.4} & \textbf{28.0} \\
                          & Qwen3-8B & 66 & 402 & 8 & 27 & \textbf{14.7} & 27.3 & 84.1 & 1.8 \\
\midrule
\multicolumn{10}{l}{\textit{Compact strategies}} \\
SheetCompressor           & GPT-5.2 & 263 & 221 & 12 & 7 & \textbf{54.7} & \textbf{64.8} & \textbf{98.2} & \textbf{25.6} \\
                          & Sonnet-4-6 & 243 & 236 & 12 & 12 & \textbf{50.7} & \textbf{63.7} & 89.9 & 23.9 \\
                          & Qwen3-8B & 46 & 407 & 11 & 39 & 11.3 & 25.4 & \textbf{81.5} & \textbf{2.6} \\
\addlinespace[1pt]
MiniSheet                 & GPT-5.2 & 199 & 280 & 14 & 10 & 42.3 & 54.3 & 84.7 & 20.7 \\
                          & Sonnet-4-6 & 230 & 248 & 20 & 5 & 49.7 & 62.8 & \textbf{92.8} & \textbf{26.2} \\
                          & Qwen3-8B & 53 & 417 & 10 & 23 & \textbf{12.5} & \textbf{26.5} & 79.3 & 2.0 \\
\bottomrule
\end{tabular}
\par\smallskip
\footnotesize Counts are abstain-aware outcomes. Acc is Pass@1, $(TP+TN)/N$. Range Overlap is expected-reference recall over expanded cell sets. Syntax and Sketch are abstain-aware: correct abstentions contribute as positives and wrong abstentions as zeros; Error rows are excluded from auxiliary metric denominators. Values in metric columns are percentages; bold marks the best value within each strategy group and model.
\end{table*}

\begin{figure}
    \centering
    \includegraphics[width=\linewidth]{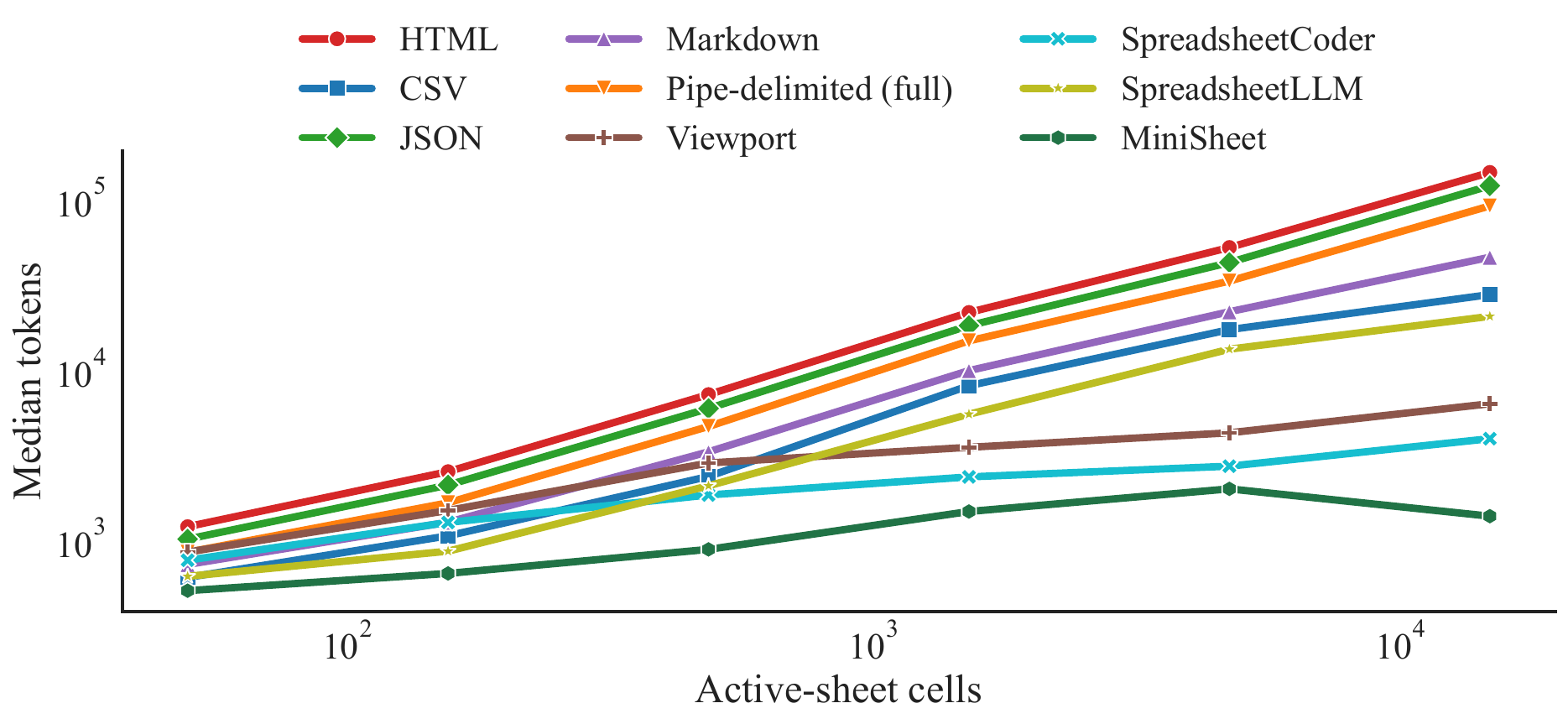}
    \caption{Token Usage}
    \label{fig:token-comparison-full}
\end{figure}

\end{document}